\documentclass{article}

    \PassOptionsToPackage{numbers, compress}{natbib}

\usepackage[preprint]{neurips_2025}

\usepackage[utf8]{inputenc} 
\usepackage[T1]{fontenc}    
\usepackage{hyperref}       
\usepackage{url}            
\usepackage{booktabs}       
\usepackage{amsfonts}       
\usepackage{nicefrac}       
\usepackage{microtype}      
\usepackage{xcolor}         

\usepackage{amsmath}
\usepackage{arydshln}
\usepackage{multicol}
\usepackage{multirow}
\usepackage{adjustbox}
\usepackage{amssymb}
\usepackage{xcolor}  
\usepackage{dsfont}
\usepackage[most]{tcolorbox}
\usepackage{subcaption}
\usepackage{caption}
\usepackage{xcolor}
\usepackage{enumitem}

\definecolor{myblue}{HTML}{4E84C4}
\definecolor{myred}{HTML}{B02418}
\definecolor{mygreen}{HTML}{34692E}
\definecolor{myorange}{HTML}{DA7842}
\definecolor{paperblue}{HTML}{077dea}
\definecolor{babyblue}{HTML}{E3EDF7} 

\newcommand{\coloredalpha}{\textcolor{paperblue}{\alpha}}
\newcommand{\coloredbeta}{\textcolor{paperblue}{\beta}}

\newcommand{\coloredsigma}{\textcolor{paperblue}{\sigma}}
\newcommand{\coloreddelta}{\textcolor{paperblue}{\delta}}
\newcommand{\coloredmu}{\textcolor{paperblue}{\mu}}

\NewDocumentCommand{\hongru}
{ mO{} }{\textcolor{red}{\textsuperscript{\textit{Hongru}}\textsf{\textbf{\small[#1]}}}}

\NewDocumentCommand{\cheng}
{ mO{} }{\textcolor{orange}{\textsuperscript{\textit{Cheng}}\textsf{\textbf{\small[#1]}}}}

\NewDocumentCommand{\xiusi}
{ mO{} }{\textcolor{cyan}{\textsuperscript{\textit{Xiusi}}\textsf{\textbf{\small[#1]}}}}

\title{Acting Less is Reasoning More ! \\ Teaching Model to Act Efficiently}

\author{%
  Hongru Wang$^{\coloredalpha}$, Cheng Qian$^{\coloredbeta}$, Wanjun Zhong$^{\coloreddelta}$, Xiusi Chen$^{\coloredbeta}$, Jiahao Qiu$^{\coloredsigma}$,  \\ 
  \textbf{Shijue Huang}$^{\coloredmu}$, \textbf{Bowen Jin}$^{\coloredbeta}$, \textbf{Mengdi Wang}$^{\coloredsigma}$, \textbf{Kam-Fai Wong}$^{\coloredalpha}$, \textbf{Heng Ji}$^{\coloredbeta}$ \\
  $^{\coloredalpha}$The Chinese University of Hong Kong, $^{\coloredbeta}$University of Illinois Urbana-Champaign \\
  $^{\coloredsigma}$Princeton University,
  $^{\coloreddelta}$Sun Yat-sen University,
  $^{\coloredmu}$Hong Kong University of Science and Technology \\
  \texttt{hrwang, kfwong@se.cuhk.edu.hk, hengji@illinois.edu} \\
}

\begin{document}

\maketitle

\begin{abstract}
Tool-integrated reasoning (TIR) augments large language models (LLMs) with the ability to invoke external tools during long-form reasoning, such as search engines and code interpreters, to solve tasks beyond the capabilities of internal reasoning.
While reinforcement learning (RL) has shown promise in training such agents, most of existing approaches typically optimize only for final correctness without considering the efficiency or necessity of external tool use. This often leads to excessive tool calling, incurring high computational costs and hindering the development of internal reasoning capabilities - a phenomenon known as \textit{cognitive offloading}. To this end, we propose Optimal Tool Call-controlled Policy Optimization (OTC-PO), a simple yet effective RL-based framework that encourages models to produce accurate answers with minimal tool calls. Our method introduces a tool-integrated reward that jointly considers answer correctness and corresponding tool use behavior of model to reach that answer. To validate the effectiveness, we introduce the metric of \textit{tool productivity}, defined as the ratio between the number of correct answers and the total number of tool calls across all test cases. This metric reflects how efficiently tool usage contributes to successful task completion, with higher values indicating smarter and more autonomous reasoning. We instantiate this framework within both Proximal Policy Optimization (PPO) and Group Relative Preference Optimization (GRPO), resulting in OTC-PPO and OTC-GRPO. Experiments with Qwen-2.5 and Qwen-Math across multiple QA benchmarks show that our approach reduces tool calls by up to 68.3\% and improves tool productivity by up to 215.4\%, while maintaining comparable answer accuracy.
\end{abstract}



\section{Introduction}
\label{sec:intro}

\begin{quote}
\textit{"The autonomous machine intelligence is designed to minimize the number of actions a system needs to take in the real world to learn a task."}  ― Yann LeCun~\citep{lecun2022path}
\end{quote}

Recent advances in large language models (LLMs) have demonstrated impressive reasoning capabilities when fine-tuned via reinforcement learning (RL) with simple rule-based rewards.
This has led to the emergence of large reasoning models (LRMs) - such as OpenAI's o1~\citep{openai2024openaio1card}, DeepSeek-R1~\citep{deepseekai2025deepseekr1}, and QwQ~\citep{qwq32b} - which exhibit creative and diverse behaviors in extended Chain-of-Thought (CoT) reasoning~\citep{wei2023chainofthoughtpromptingelicitsreasoning}.
These models develop meta-reasoning capabilities reminiscent of human cognition, such as reflection and decomposition, which have been shown to enhance performance on complex tasks~\citep{wang2025selfreasoning, qin2024o1replicationjourneystrategic, zeng2025simplerlzooinvestigatingtamingzero}.
However, the internal reasoning capacity of LRMs alone is often insufficient or even infeasible for tasks that require interaction with external environments, such as accessing up-to-date domain knowledge or performing precise computations. To overcome these limitations, tool-integrated reasoning (TIR) has emerged as a promising paradigm~\citep{gou2024tora, li2025startselftaughtreasonertools, wang-etal-2023-large, qian2025smartselfawareagenttool, tool_learning_tut}, enabling LLMs to interface with external tools like search engines, code interpreters and even physical robots. By bridging language understanding with external capabilities, TIR significantly expands the problem-solving scope of LLMs beyond internal knowledge-only reasoning.

\begin{figure}
    \centering
    \includegraphics[trim={1cm 0cm 0cm 0cm}, clip, width=1.0\textwidth]{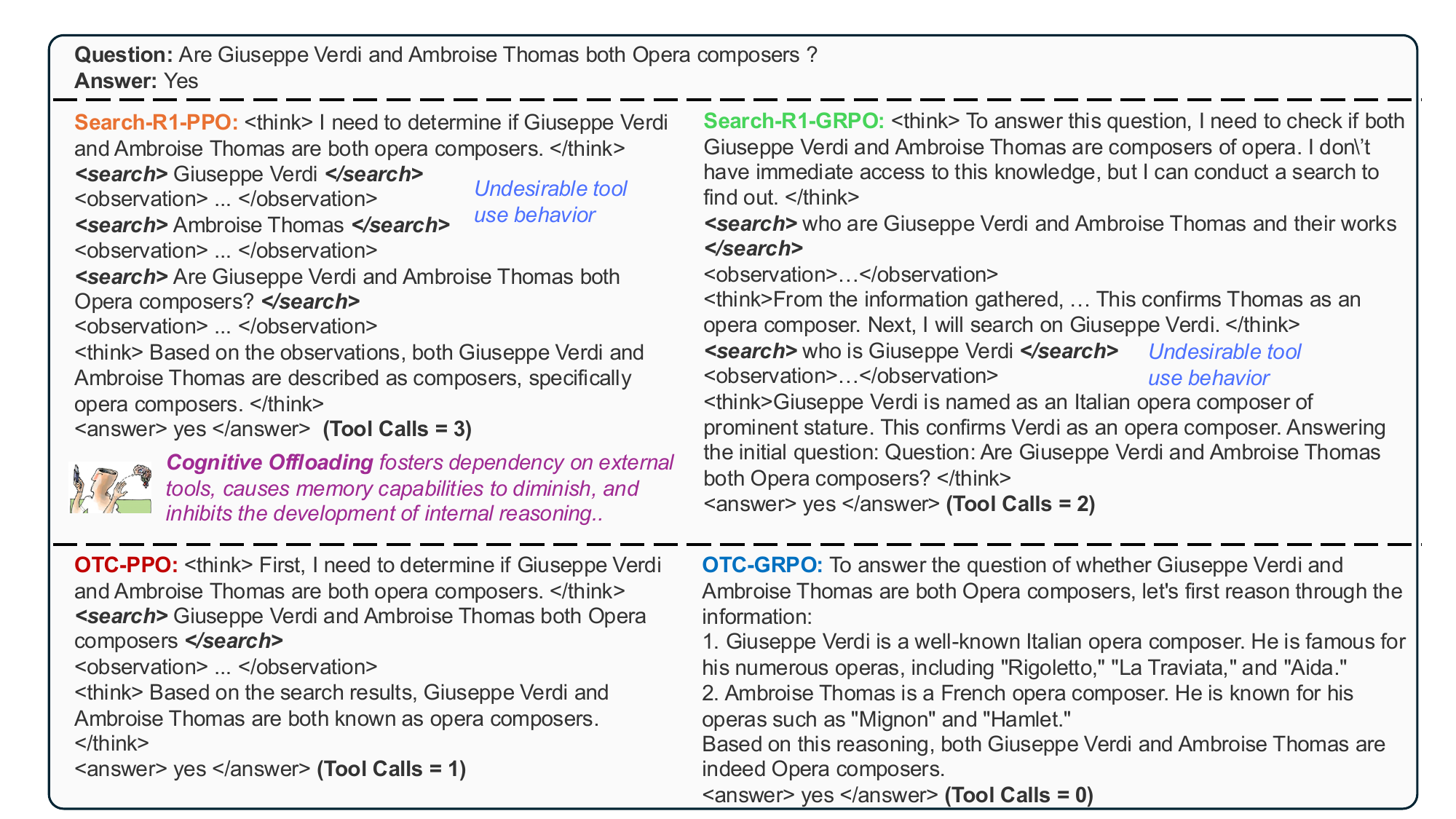}
\caption{A case study using \textit{search} as a tool on Qwen-7B-Base model, where all four methods arrive at the correct answer but exhibit distinct behaviors in internal reasoning and external tool usage.}
\label{fig:main_case_study}
\vspace{-4mm}
\end{figure}

While recent efforts have sought to extend RL to TIR by leveraging rule-based rewards tied to final answer correctness~\citep{jin2025searchr1trainingllmsreason, li2025torlscalingtoolintegratedrl, chen2025researchlearningreasonsearch}, these approaches fall short in modeling the optimal tool use behavior of model to arrive at such answer. Despite achieving strong performance and improved generalization, current methods often encourage indiscriminate tool usage, introducing critical challenges during both training and inference. On the one hand, frequent and unnecessary tool calls incur substantial computational and temporal overhead, as executing tools - especially at scale - demands significant infrastructure and increases operational costs, limiting deployment feasibility. On the other hand, extensively rely on external tools may hinder the development and utilization of its own internal reasoning capabilities — a phenomenon commonly known as \textit{cognitive offloading}~\citep{risko2016cognitive}. Fig.~\ref{fig:main_case_study} presents a representative example. The baseline method (i.e., Search-R1~\citep{jin2025searchr1trainingllmsreason}), which optimizes only for final answer accuracy, excessively invokes external tools, even when the model is capable of solving the problem independently without any tools. By offloading reasoning to external tools, the model bypasses opportunities to strengthen its internal reasoning abilities, thereby ultimately undermining long-term generalization and autonomy \footnote{Our empirical study shows that this issue becomes more severe as the model size increases.}.


Therefore, we argue that a truly autonomous agentic LLM should learn to produce correct answers or complete tasks successfully with the minimal number of external tool calls - an objective that aligns closely with Yann LeCun’s vision for autonomous machine intelligence~\citep{lecun2022path}. We aim to develop such model based on a foundational assumption: for \textit{each question} and \textit{each model}, there exists an optimal number of tool calls, defined as the minimal number required for the model to arrive at a correct answer. Crucially, this optimal number could be varied across models or questions. For the same question, models with different reasoning or representation capabilities may require different levels of tool assistance. Similarly, even for a single model, the number of necessary tool calls can vary significantly depending on the complexity of different question. For example, some questions may be solvable through internal reasoning alone - requiring zero tool calls\footnote{In such cases, the task reduces to pure internal knowledge-only reasoning.} - while others may necessitate one or more tool interactions to yield a correct answer. This variability also highlights the
unique advantages of RL in steering the behavior of an agent on a per-question, per-model basis, rather than a just one-fits-all method (i.e., SFT). Ideally, among all tool-integrated reasoning trajectories that lead to the correct answer for the same question, the reward function in RL should assign the highest reward to the trajectory that uses the optimal (i.e., minimal) number of tool calls. Other correct trajectories that involve excessive tool calls - by definition, suboptimal - should be rewarded less \footnote{All such trajectories incur unnecessary tool usage and are thus less efficient.}. This formulation allows the model to learn policies that jointly optimize for both answer correctness and optimal tool use behavior on a per-question, per-model basis.

We then propose Optimal Tool Calls controlled Policy Optimization (OTC-PO), a simple yet effective RL-based method that enables models to learn optimal tool use behavior besides providing correct answer.
OTC-PO trains models to minimize the number of tool calls required to reach a correct solution, thereby reducing both training and inference costs without sacrificing accuracy.
To achieve this, we introduce a tool-integrated reward that modulates traditional reward signals, such as correctness, with a scaling \textit{coefficient} reflecting tool efficiency. This encourages the model to prioritize correct answers that require fewer tool invocations.
Our formulation shifts the optimization objective from correctness alone to tool productivity, defined as the ratio between task benefit (e.g., answer accuracy) and tool usage cost (e.g., number of tool calls). 
To the best of our knowledge, this is the pioneering RL-based framework that explicitly optimizes for both the efficiency and effectiveness of tool-integrated reasoning.
OTC-PO is lightweight and plug-and-play, requiring only minimal changes to standard RL pipelines, making it easy to adopt in existing systems.
We conduct comprehensive experiments on two widely used tools: search and code based on several LLMs, Qwen-2.5-3B/7B-Base and Qwen2.5-Math-1.5B/7B-Base~\cite{qwen2025qwen25technicalreport}, and derive models with a substantial tool calls reduction in trajectory while approximately maintaining accuracy with strong baselines. In summary, the key contributions are as follows.

\begin{itemize}[leftmargin=*,topsep=1pt,itemsep=1pt]
    \item We are the pioneering study to i) consider tool use behavior beyond final answer correctness by proposing a new task formulation that has been largely overlooked in prior work; ii) identify the cognitive offloading phenomenon in TIR of LLMs; iii) introduce the concept of tool productivity to measure the effectiveness and efficiency of TIR.

    \item We propose a simple, faster, scalable, and generalizable OTC-PO algorithm to encourage the model to use optimal tool calls to solve the problem and maximize tool productivity, based on the fundamental observation that there exists a minimal number of tool calls required for each question and model pair. We note that it is compatible with various RL algorithms and can be easily implemented. 

    \item We implement OTC-PPO and OTC-GRPO as two typical methods without losing the adaptability and generalization based on PPO~\cite{schulman2017proximal} and GRPO~\cite{shao2024deepseekmath} algorithms, and the experimental results on several benchmarks and baselines demonstrate significant reductions in tool call cost while preserving most of the accuracy in both in-domain and out-of-domain evaluation.
\end{itemize}



\section{Related Work}

\paragraph{Tool Utilization for LLMs.} Teaching LLMs to use tools enables them to interact with external environments while overcoming several inherent limitations such as restricted access to up-to-date or domain-specific knowledge and poor mathmatical operation capabilities. There are three major methods which can achieve this goal: 1) prompting engineering~\citep{qian-etal-2023-creator, wang-etal-2024-appbench, qian2024investigate, yuan2024craftcustomizingllmscreating, qian2024escapebench, wang2025selfdcreasonactself}, which guides the model’s behavior through carefully designed input templates or few-shot examples without modifying model weights; 2) supervised finetuning on tool-integrated reasoning datasets~\citep{qin2023toolllmfacilitatinglargelanguage,gou2024tora,qian2024toolink,chen2025advancingtoolaugmentedlargelanguage,qian2025smartselfawareagenttool}, where the model learns from annotated trajectories that demonstrate when and how to use tools in context; and 3) reinforcement learning~\citep{jin2025searchr1trainingllmsreason, li2025torlscalingtoolintegratedrl, feng2025retoolreinforcementlearningstrategic, qian2025toolrlrewardisalltool}, which allows the model to directly learn tool-use strategies by interacting with an environment and optimizing long-term rewards, enabling more adaptive and goal-directed behaviors. While prompting and supervised fine-tuning have shown promising results, they rely heavily on expert-designed prompts and tool-integrated annotations, which limits their scalability and generalization. As a result, recent efforts have shifted toward reinforcement learning, which demonstrates improved performance and generalization through simple rule-based rewards.

\paragraph{Tool-integrated Reasoning Efficiency.} Only few of studies from prompting engineering and supervised fine-tuning attention on tool-integrated reasoning efficiency issues in terms of the cost of tool usages during the reasoning ~\citep{wang2025selfdcreasonactself, shen-etal-2024-smartcal, qian2025smartselfawareagenttool}. In detail, Wang et al. \citep{wang2025selfdcreasonactself} first propose a prompting-based framework: Self Divide-and-Conquer (SelfDC) to leverage the self-aware confidence score of LLMs to decide whether or not need to call tools during reasoning, achieving better trade-off between effectiveness and efficiency in the context of RAG. Furthermore, several works follow this direction and explore more boarder applications and larger tool spaces ~\citep{qian2025smartselfawareagenttool, shen-etal-2024-smartcal, li2025adaptivetooluselarge}. For example, SMART~\citep{qian2025smartselfawareagenttool} collect the well-designed dataset to finetune the model to only call tools when the knowledge is outside the inherent parametric knowledge of LLMs. Despite these advancements, most existing approaches still rely on complex prompt engineering or annotated datasets, which hinders their adaptability and scalability to new scenarios. In contrast, the efficiency of tool-integrated reasoning within reinforcement learning frameworks remains largely underexplored.







\section{Methodology}

In this section, we first provide a definition of task considering the both effectiveness and efficiency of tool-integrated reasoning, followed by general RL framework and then our proposed Optimal Tool Call-controlled Policy Optimization (OTC-PO).

\subsection{Task Definition}

Given a question $q$, and an environment $\mathcal{E}$ that provides access to a set of tools $\mathcal{T}= \{t_0, t_1, ...t_n\}$, the language model $\mathcal{M}$ can \textit{optionally} interact with the environment by calling specific tools in $\mathcal{T}$, obtaining the corresponding tool results from $\mathcal{E}$, and iteratively repeating this processing until the final answer is driven. Without losing generalization, the tool-integrated reasoning trajectory $\tau_k$ at step $k$ is defined as follows:

\vspace{-4mm}
\begin{equation}
    \tau_k = (r_1, tc_1, o_1), (r_2, tc_2, o_2), ... (r_k, tc_k, o_k),
\end{equation}

where $r_i, tc_i, o_i$ denotes the reasoning, tool call and returned observation respectively. Importantly, we also account for reasoning steps that do not involve tool usage. Suppose the step $p$ does not need to call tools, then the $tc_p$ and $o_p$ become empty string, the reasoning content $r_p$ can either be merged with the subsequent reasoning step $r_{p+1}$ to form the new $r^*_{p+1}$, or, if $p$ is the last step \footnote{The last step is practically determined by the predefined maximal tool calls.}, be used directly to derive the final answer. The objective of the task is to generate the correct answer $\hat{a}$ with the minimal cost of the full trajectory $\tau$ as follows:

\vspace{-3mm}
\begin{equation}
    \mathop{\arg\min}_{\tau} \,\, \text{Cost}(\tau) \quad \text{subject to} \quad \mathcal{M}(q, \tau) = \hat{a},
\end{equation}

Here the cost is measured as the number of tool calls within the trajectory $\tau$ considering its simplicity and generalization. Thus the model is encouraged to not only generate correct answer but also minimize the cost. We emphasize that this revised objective is not merely a minor adjustment, but a paradigm shift, redirecting focus from answer correctness alone to the broader behavioral efficiency of tool use during reasoning.

\subsection{Tool-integrated Reinforcement Learning}

Inspired by recent success to use RL for better reasoning in LLMs, several efforts try to extend RL to tool-integrated reasoning with the objective functions can be defined as follows:

\vspace{-2mm}
\begin{equation}
    \max_{\pi_\theta} \mathbb{E}_{q \sim \mathcal{D},\, y \sim \pi_\theta(\cdot \mid q; \mathcal{E})} \left[ r_\phi(q, y) \right] - \beta \, \mathbb{D}_{kl} \left[ \pi_\theta(y \mid q; \mathcal{E}) \,\|\, \pi_{\mathrm{ref}}(y \mid q; \mathcal{E}) \right],
\end{equation}

where $\pi_\theta$, $\pi_{ref}$ stand for the policy model and reference model respectively, $r_\phi$ is the reward function and $\mathbb{D}_{kl}$ is the KL-divergence measure. $q$ is the question drawn from the dataset $\mathcal{D}$ and $y$ is the generate outputs consists of the tool-integrated reasoning trajectory $\tau$ and the final answer $a$. To optimize this goal, there are two well-established policy-gradient RL methods: Proximal Policy Optimization (PPO) and Group Relative Policy Optimization
(GRPO) in TIR. We provide detailed optimization objective in Appendix~\ref{appendix_trl} to save space.

\paragraph{Reward Design.}

Reward signals serve as the optimization objective and directly guide the behavior of the policy model during training. Most prior predominantly focuses on rule-based correctness rewards ~\citep{jin2025searchr1trainingllmsreason, li2025torlscalingtoolintegratedrl}, as illustrated below:

\vspace{-3mm}
\begin{equation}
    r_\phi (q, y) = r_{correct} = 1 \text{ if } a = \hat{a} \text{ else } 0,
\end{equation}

where $a$ is the extracted predicted answer from the response $y$ and $\hat{a}$ is the ground truth answer. Some of them further consider the format reward by assigning additional reward besides the correctness such as $r_{format}$ if the model generate the response in the required format. We discuss this in Appendix.

\begin{figure}
    \centering
    \includegraphics[trim={2cm 7cm 1cm 3cm}, clip, width=1.0\textwidth]{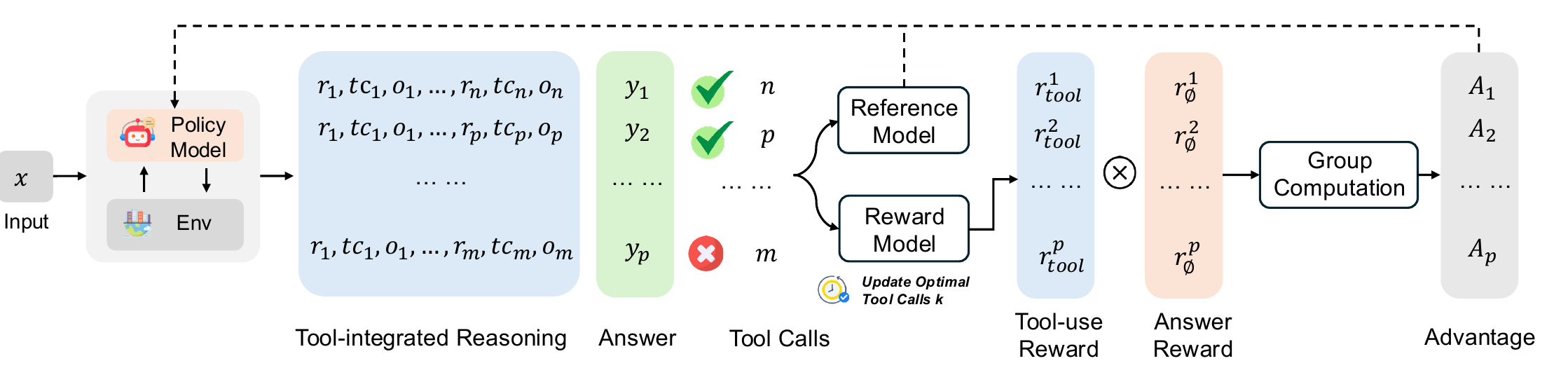}
    \caption{An overview of OTC-GRPO Algorithm.}
    \label{fig:otc_grpo}
    \vspace{-6mm}
\end{figure}

\subsection{OTC-PO: Optimal Tool Call-controlled Policy Optimization}

Instead of only considering the correctness (and optionally format) of the answer, we argue it is important to consider the effectiveness and efficiency of tool-integrated reasoning trajectory. Specifically, different questions may require different numbers of tool interactions depending on the internal capabilities of  model, and for each question–model pair, there exists an optimal (i.e., minimal) number of tool calls needed to reach the correct answer. Given the real-world cost and latency associated with tool calls, it is crucial to move beyond coarse, answer-only rewards and instead provide finer-grained signals that capture the overall quality of TIR trajectory.

To this end, we propose Optimal Tool Call-controlled Policy Optimization (OTC-PO), can be easily integrated with any RL algorithms, and is broadly applicable across different models and settings, aiming to preserve overall accuracy while significantly reducing the cost of tool interactions. At the heart of OTC-PO is a reward mechanism that ranks the current number of tool calls $m$ relative to the optimal number of tool calls $n$ given the question and model. In most realistic settings, the true optimal number $n$ is unknown in advance. To address this, OTC-PO approximates $n$ by tracking the minimal number of tool calls observed across different trajectories to arrive the correct answer for the same question. This approach enables the model to learn efficient tool use without requiring prior knowledge of the optimal tool budget. Nonetheless, the framework can be naturally extended to scenarios where the optimal tool call number is known a priori (e.g., $n = 0$ for language-only reasoning). We instantiate OTC-PO in two variants: OTC-PPO and OTC-GRPO, tailored to different underlying RL algorithms.



\paragraph{OTC-PPO.} Since there is no multiple trajectories for same question $q$ in single optimization step, we design the tool reward according to the number of tool calls $m$ in the current trajectory as follows:

\vspace{-4mm}
\begin{equation}
\label{ppo_tool_reward}
r_{tool} = \cos(\frac{m * \pi}{2m + c})
\end{equation}

The core idea is to map $m$ to [0, $\pi$/2) and $c$ is the smooth constant that controls the reward decay rate. The smaller the value, the faster the punishment and the more it encourages less use of tools; the larger the value, the more tolerant it is. In this way, among trajectories, those requiring more tool calls will receive lower rewards than those achieved with fewer tool calls. Although we can track the $m$ here for approximation of $n$ in later steps, we empirically find it requires much longer training steps due to poor sampling efficiency in PPO.


\paragraph{OTC-GRPO.} We first identify the trajectories $\{\tau^1, \tau^2, ..., \tau^p \}$ leading to correct answer from the group for the question $q$, and get the corresponding number of tool calls for each trajectory: $\mathcal{C}$ = $\{c^1, c^2, ..., c^p\}$, and we can calculate the minimal tool calls $k$ = $\min(\mathcal{C})$, serving as the \textit{local} approximation of optimal tool calls for $q$. Furthermore, we can update $k$ during multiple epochs to approximate the \textit{global} optimal tool calls if the policy model finds the better solution with less than $k$ calls in later iterations. We use $n$ to indicate the approximation of optimal tool calls, and $m$ the number of tool calls in the current trajectory for $q$. Therefore we design the reward as follows:

\vspace{-2mm}
\begin{equation}
r_{tool} = 
\begin{cases}
1 & \text{if } f(m, n) = n = 0   \\
\cos(\frac{m * \pi}{2m + c}) &\text{if } n = 0 \\
\sin (\frac{f(m, n) * \pi}{2n}) & \text{otherwise} \\
\end{cases} \quad
f(m, n) = 
\begin{cases}
0, & \text{if } m = 0 \text{ and } n = 0 \\
m, & \text{if } n = 0 \\
\dfrac{2nm}{m + n} , & \text{otherwise}
\end{cases}
\end{equation}


where $f(m,n)$ is the mapping function to re-map the $m$ to the range from 0 to $2n$. Briefly, the key here is to assign the highest reward 1 (a.k.a, $\sin(\pi/2)$) when the policy model achieves optimal tool calls $n$, and when actual number of tool calls $m$ deviates from $n$, either exceeding or falling short, the model receives a comparatively lower reward \footnote{We provide the reward illustration in the Appendix for better understanding.}. Thus, we can assign different rewards dynamically according to the different $n$ and $m$ for the question $q$. 

\paragraph{Tool-integrated Reward Design.} Inspired by recent study \citep{arora2025traininglanguagemodelsreason}, we regard the tool-integrated reward $r_{tool}$ as a \textit{coefficient} of conventional reward function $r_\phi (q, y)$, leading to the final tool-integrated reward function:

\vspace{-4mm}
\begin{equation}
\label{eq:tool_total_reward}
    r^{tool}_\phi(q,y) = \alpha * r_{tool} * r_\phi(q, y)
\end{equation}

where $\alpha$ is a hyperparameter that controls the scaling of the tool reward as $r_{\phi}$ is usually same for correct or wrong answer. Notably, this multiplicative structure ensures that tool efficiency is mainly rewarded when the primary task objective (e.g., answer correctness) is satisfied. For example, when the final answer is incorrect, ${r_\phi}$ is 0, effectively disabling the influence of $r_{tool}$. This design helps mitigate reward hacking by preventing the policy model from being incentivized to use tools without ultimately producing a correct answer. If the answer is correct, $r_\phi$ is 1 and then \(r_{tool}\) begins to dominate the scale of the reward, thus encouraging less tool usage. We highlight this design is non-trivial since there are several advantages of this design: i) It maximally preserves overall accuracy, with a theoretical justification~\citep{arora2025traininglanguagemodelsreason}; ii) It mitigates the risk of reward hacking compared to additive forms (i.e., $r_{tool}+r_{\phi}$), which we found to be unstable and sub-optimal \footnote{We empirically found that the model achieves higher reward by call less tools without correct answer if we use addition operation.}; iii) It generalizes well to different formulations of $r_\phi$, such as $r_\phi = r_{correct}$ or $r_\phi=r_{correct} + r_{format}$. We provide the theoretical justification, more explanations and examples in the Appendix~\ref{appendix:reward}.


\section{Experiments}
\label{sec:exp}

\subsection{Set up}

\paragraph{Datasets and Baselines.} We mainly follow the Search-R1~\citep{jin2025searchr1trainingllmsreason} and ToRL~\citep{li2025torlscalingtoolintegratedrl} setting and use same baselines and datasets for the fair and comprehensive evaluation. Specifically, we use NQ and HotpotQA as training dataset for search, and we use the dataset provided in ToRL for code. We also directly compare our methods with several baselines such as SFT, Base-RL, retrieval-augmented generation baselines (i.e., RAG, IRCoT) and more importantly the Search-R1 and ToRL themselves.

\paragraph{Evaluation Metrics.} Besides the exact match (EM) and the average tool calls (TC), we additionally define a new metric, \textit{tool productivity} (TP), which measures the effectiveness and efficiency of tool calls during inference. Specifically, TP is defined as the number of correctly answered questions per unit of tool call: TP = $\frac{\sum_{i=1}^{N} \mathbb{I}\{y_i = \hat{y}_i\}}{\sum_{i=1}^{N} tc_i}$ \footnote{This can also be understood as the fraction between benefits and cost.}, where $\mathbb{I}$ is the indicator function that equals 1 if the predicted answer $\hat{y}_i$ matches the ground truth $y_i$, and $tc_i$ denotes the number of tool calls used in the $i_{th}$ instance. This metric reflects how efficiently the model converts tool usage into correct answers, capturing both utility and cost in a single measure. While EM provides a standard measure of accuracy, it does not reflect the underlying reasoning cost. Therefore, we consider TP as more informative indicators of agentic reasoning efficiency -- highlighting not only whether the model can provide correct answer, but how economically it arrives at that correctness. 

\paragraph{Implementation Details.} We re-use the same parameter in Search-R1~\citep{jin2025searchr1trainingllmsreason} and ToRL~\citep{li2025torlscalingtoolintegratedrl} respectively. There are only minor modification we make to suit our method: i) We slightly change the template in Search-R1 to tell the model that it only need to call tolls when necessary (Appendix), and we do not change the template in ToRL; ii) We set the maximal number of tool calls $\mathcal{C}$ in ToRL to 3 to better study the effects of our methods when multiple tool calls are allowed and keep it as 4 as in original Search-R1. Moreover, we set $c$ as corresponding max turns or maximal tool limits, $\alpha$ as 1. We conduct our experiments on 8 A100-80G GPU and re-produce the results of Search-R1 and ToRL independently. We implement OTC-GRPO using the global approximation of optimal tool calls since it leads to more stable and better optimization.

\begin{table*}[!t]
\setlength{\belowcaptionskip}{0pt}
    \centering
    \caption{The results of OTC-PO with different baselines in search. The results except Search-R1 are directly copied from original paper~\citep{jin2025searchr1trainingllmsreason}. We highlight the relative improvement of OTC-PO and OTC-GRPO compared with the corresponding baseline Search-R1-PPO and Search-R1-GRPO. It is found that our method can achieve comparable performance with much lower TC and higher TP.}
    \begin{adjustbox}{max width=0.9 \textwidth}
    \begin{tabular}{l| ccc|ccc}
    \toprule
    \multirow{2}{*}{\textbf{Models}} & \multicolumn{3}{c}{\textbf{NQ}} & \multicolumn{3}{|c}{\textbf{HotpotQA}} \\  
    \cline{2-7} & EM ($\uparrow$) & TC ($\downarrow$) & TP ($\uparrow$) & EM ($\uparrow$) & TC ($\downarrow$) & TP ($\uparrow$) \\
    \hline
    \multicolumn{3}{l}{\textbf{Qwen2.5-3B(-Base)}} \\
    R1-Base & 0.226 & - & - & 0.201 & - & - \\
    SFT & 0.249 & - & - & 0.186 & - & - \\
    RAG & 0.348 & 1.0 & 0.348 & 0.255 & 1.0 & 0.255\\
    IRCoT & 0.111 & 10.0 & 0.011 & 0.164 & 10.0 & 0.016 \\
    Search-R1-PPO & 0.403 & 1.738 & 0.232 & 0.279 & 1.716 & 0.163 \\
    Search-R1-GRPO & 0.404 & 1.426 & 0.283 & 0.312 & 1.802 & 0.173 \\

    \hdashline
    OTC-PPO & 0.355 & 1.010 (\textcolor{green!60}{$\blacktriangledown$ 41.9\%}) & 0.351 (\textcolor{red!60}{$\blacktriangle$ 51.3\%}) & 0.260 & 1.026 (\textcolor{green!60}{$\blacktriangledown$ 40.2\%}) & 0.253 (\textcolor{red!60}{$\blacktriangle$ 55.2\%}) \\
    OTC-GRPO & 0.444 & 1.008 (\textcolor{green!60}{$\blacktriangledown$ 29.3\%}) & 0.440 (\textcolor{red!60}{$\blacktriangle$ 55.5\%}) & 0.365 & 1.387 (\textcolor{green!60}{$\blacktriangledown$ 23.0\%}) & 0.263 (\textcolor{red!60}{$\blacktriangle$ 52.0\%})  \\

    \hline
    \multicolumn{3}{l}{\textbf{Qwen2.5-7B(-Base)}} \\
    R1-Base & 0.270 & - & - & 0.242 & - & - \\
    SFT & 0.318 & - & - & 0.217 & - & - \\
    RAG & 0.349 & 1.0 & 0.349 & 0.299 & 1.0 & 0.299 \\
    IRCoT & 0.224 & 9.999 & 0.022 & 0.133 & 9.982 & 0.013 \\
    Search-R1-PPO & 0.449 & 3.282 & 0.136 & 0.380 & 3.741 & 0.102 \\
    Search-R1-GRPO & 0.399 & 1.697 & 0.235 & 0.341 & 2.109 & 0.162  \\

    \hdashline
    OTC-PPO & 0.446 & 1.040 (\textcolor{green!60}{$\blacktriangledown$ 68.3\%}) & 0.429 (\textcolor{red!60}{$\blacktriangle$ 215.4\%}) & 0.383 & 1.464 (\textcolor{green!60}{$\blacktriangledown$ 60.9\%}) & 0.262 (\textcolor{red!60}{$\blacktriangle$ 156.9\%}) \\
    OTC-GRPO & 0.444 & 0.990 (\textcolor{green!60}{$\blacktriangledown$ 41.7\%}) & 0.448 (\textcolor{red!60}{$\blacktriangle$ 90.6\%}) & 0.366 & 1.005 (\textcolor{green!60}{$\blacktriangledown$ 52.3\%}) & 0.364 (\textcolor{red!60}{$\blacktriangle$ 124.7\%}) \\
    
    \bottomrule
    \end{tabular}
    \end{adjustbox}
    \label{tab:main_exp_search}
\end{table*}
\begin{table*}[!t]
\setlength{\belowcaptionskip}{0pt}
    \centering
    \setlength{\tabcolsep}{5pt} 
    \caption{The results of OTC-PO with different baselines in ToRL~\cite{li2025torlscalingtoolintegratedrl}.}
    \begin{adjustbox}{max width=0.9 \textwidth}
    \begin{tabular}{l| ccc|ccc}
    \toprule
    \multirow{2}{*}{\textbf{Models}} & \multicolumn{3}{c}{\textbf{AIME24}} & \multicolumn{3}{|c}{\textbf{AIME25}} \\  
    \cline{2-7} & EM ($\uparrow$) & TC ($\downarrow$) & TP ($\uparrow$) & EM ($\uparrow$) & TC ($\downarrow$) & TP ($\uparrow$) \\
    \hline
    \multicolumn{3}{l}{\textbf{Qwen2.5-Math-1.5B(-Base)}} \\
    Qwen2.5-Math-1.5B-Ins & 10.0 & - & - & 10.0 & - & -  \\
    Qwen2.5-Math-1.5B-Ins-TIR & 13.3 & 1.1 & 12.1 & 13.3 & 1.4 & 9.5 \\
    ToRL-GRPO & 23.3 & 2.2 & 10.6 & 23.3 & 2.3 & 10.1 \\

    \hdashline
    OTC-GRPO & 20.0 & 1.1 (\textcolor{green!60}{$\blacktriangledown$ 50.0\%}) & 18.2 (\textcolor{red!60}{$\blacktriangle$ 71.7\%}) & 20.0 & 1.1 (\textcolor{green!60}{$\blacktriangledown$ 41.2\%}) & 18.2 (\textcolor{red!60}{$\blacktriangle$ 80.2\%}) \\

    \hline
    \multicolumn{3}{l}{\textbf{Qwen2.5-Math-7B(-Base)}} \\
    Qwen2.5-Math-7B-Ins & 10.0 & - & - & 16.7 & - & - \\
    Qwen2.5-Math-7B-Ins-TIR & 26.7 & 1.6 & 16.4 & 16.7 & 1.4 & 12.2 \\
    Base-RL & 33.3 & - & - & 6.7 & - & -\\
    ToRL-GRPO & 36.7 & 2.1 & 17.5 & 26.7 & 2.1 & 12.7 \\
    
    \hdashline
    
    OTC-GRPO & 36.7 & 0.7 (\textcolor{green!60}{$\blacktriangledown$ 66.7\%}) & 52.4 (\textcolor{red!60}{$\blacktriangle$ 199.4\%}) & 23.3 & 0.8 (\textcolor{green!60}{$\blacktriangledown$ 61.9\%}) & 29.1 (\textcolor{red!60}{$\blacktriangle$ 129.1\%}) \\
    \bottomrule
    \end{tabular}
    \end{adjustbox}
    \label{tab:main_exp_code}
    \vspace{-4mm}
\end{table*}

\subsection{Main Results}

\paragraph{Search as Tool.} Table~\ref{tab:main_exp_search} shows the results for search-required tasks. There are several key insights can be drawn: i) OTC-PPO achieves significant improvement in terms of TC and TP compared with Search-R1-PPO, resulting in up to a 68.3\% reduction in TC and a 215.4\% increase in TP, and OTC-GRPO can further reduce the absolute tool calls due to more accurate approximation; ii) It is found that our method will not sacrifice the accuracy too much especially for larger LLMs, as evidenced by our OTC-PO achieves comparable EM score with Search-R1 on Qwen2.5-7B model; iii) Different models have different tool use behaviors on different datasets. If we look at the Search-R1, we can find that 3B model tends to use less tool calls compared with 7B model. This reveals a critical issue: as language models scale up, they may tend to over-rely on external tool calls unless explicitly penalized for such behavior. This not only exacerbates the problem with larger models but also leads to an underutilization of their inherent reasoning capabilities (See \$~\ref{case_study}). The number of TC is also various across the datasets, which relates to many factors such as inherent capabilities of different models (i.e., self-awareness) and complexity of different datasets, leading to varying optimal number of tool calls for each question and model; iv) As the model size increases, the TC and TP get bigger boost no matter in OTC-PPO or OTC-GRPO. We highlight the ii) and iv) are very important since they compose the great potential scalability for our methods as model scales. 

\paragraph{Code as Tool.} We mainly follow the same setting and only report the GRPO results to be consistent with original ToRL~\cite{li2025torlscalingtoolintegratedrl} paper. Table~\ref{tab:main_exp_code} shows the results. We observe several similar findings: i) our method does not sacrifice the accuracy too much and even brings some improvement when the model gets larger. We attribute this to the development of internal reasoning capability of models when it is enforced to minimizing external tool calls, as evidenced in our case study; ii) our method can significantly boost the tool productivity, reducing the unnecessary tool calls; iii) As model size increases, our method is more effective to improve the tool productivity.

\section{Analysis}

\begin{figure}[t]
  \centering
  \begin{subfigure}[b]{0.32\textwidth}
    \includegraphics[width=\linewidth]{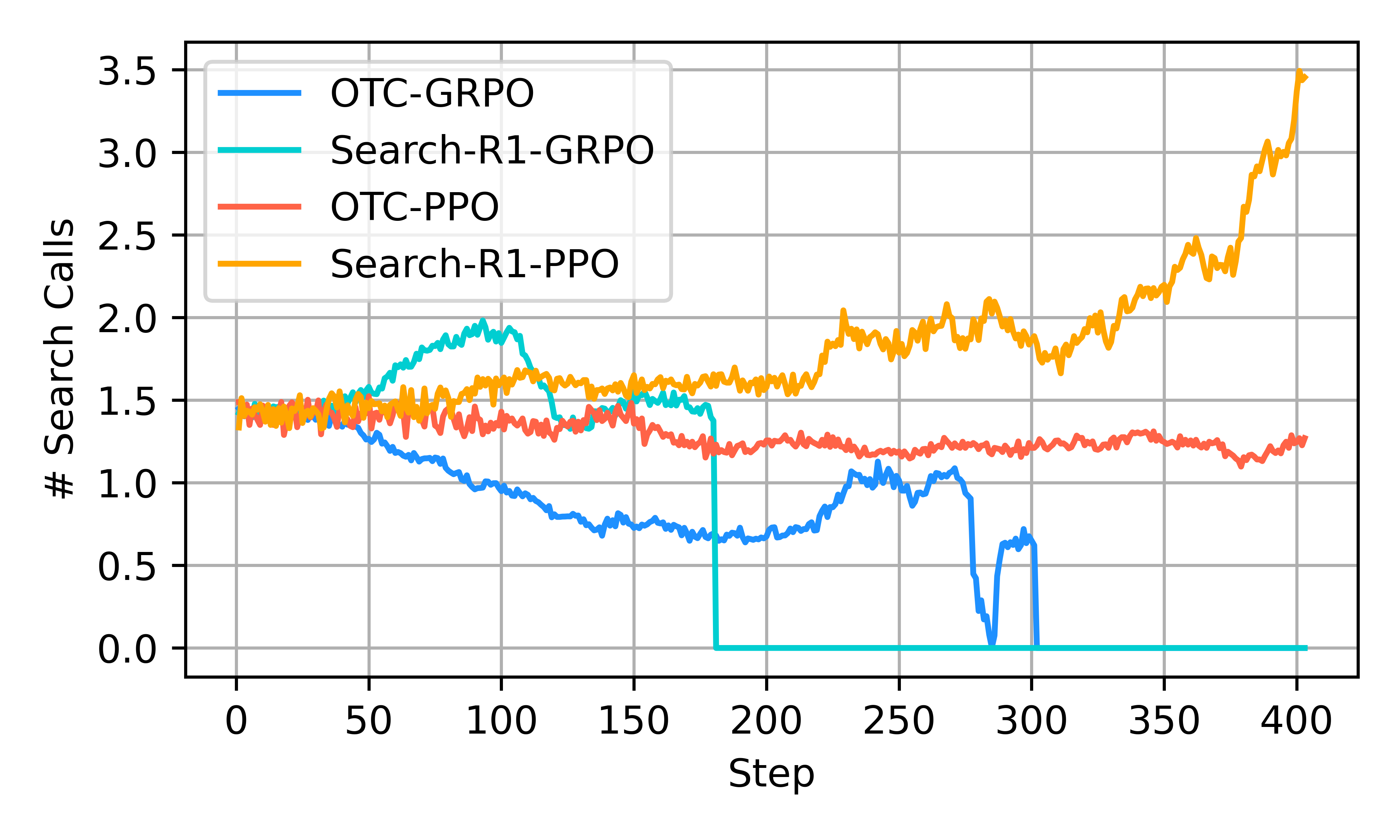}
  \end{subfigure}
  \begin{subfigure}[b]{0.32\textwidth}
    \includegraphics[width=\linewidth]{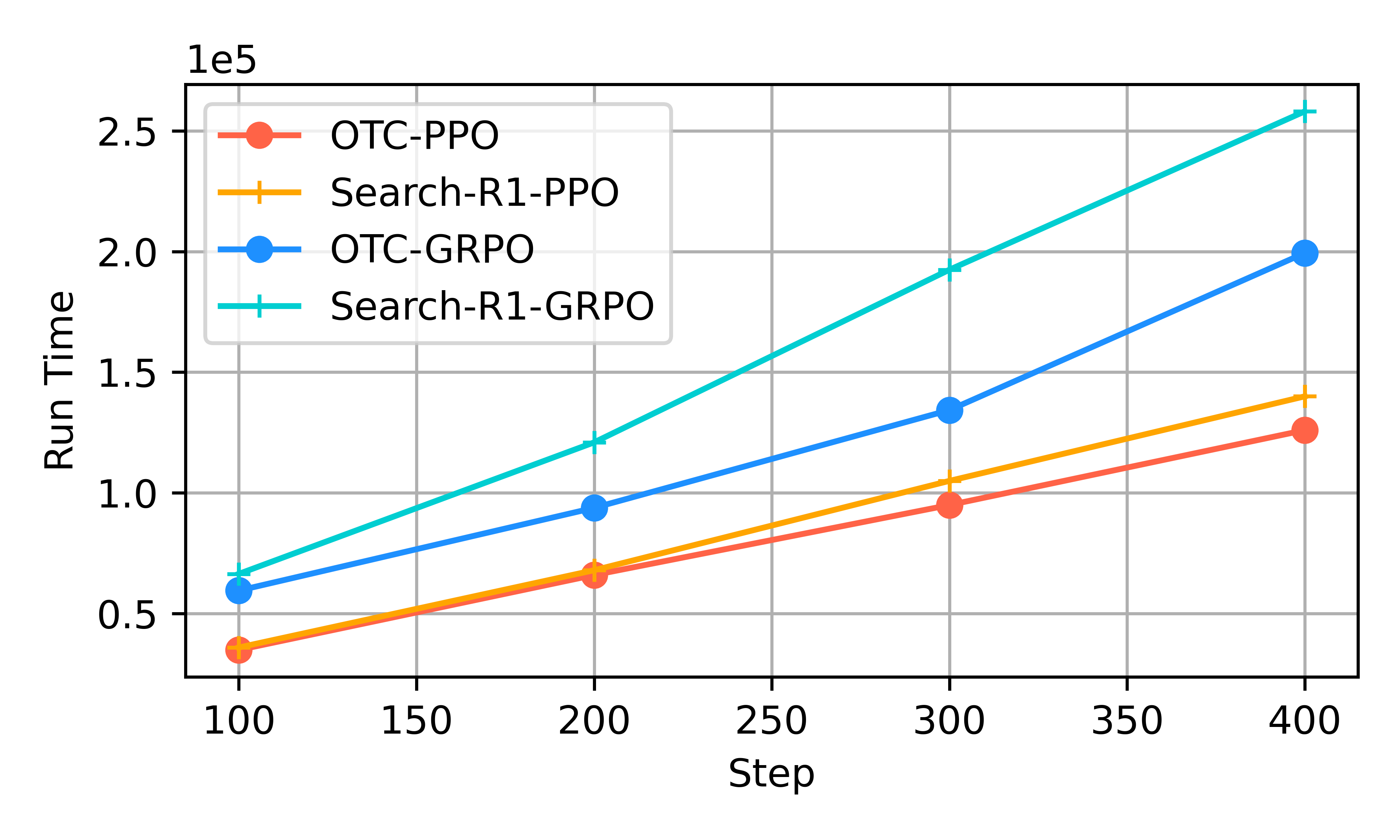}
  \end{subfigure}
  \begin{subfigure}[b]{0.32\textwidth}
    \includegraphics[width=\linewidth]{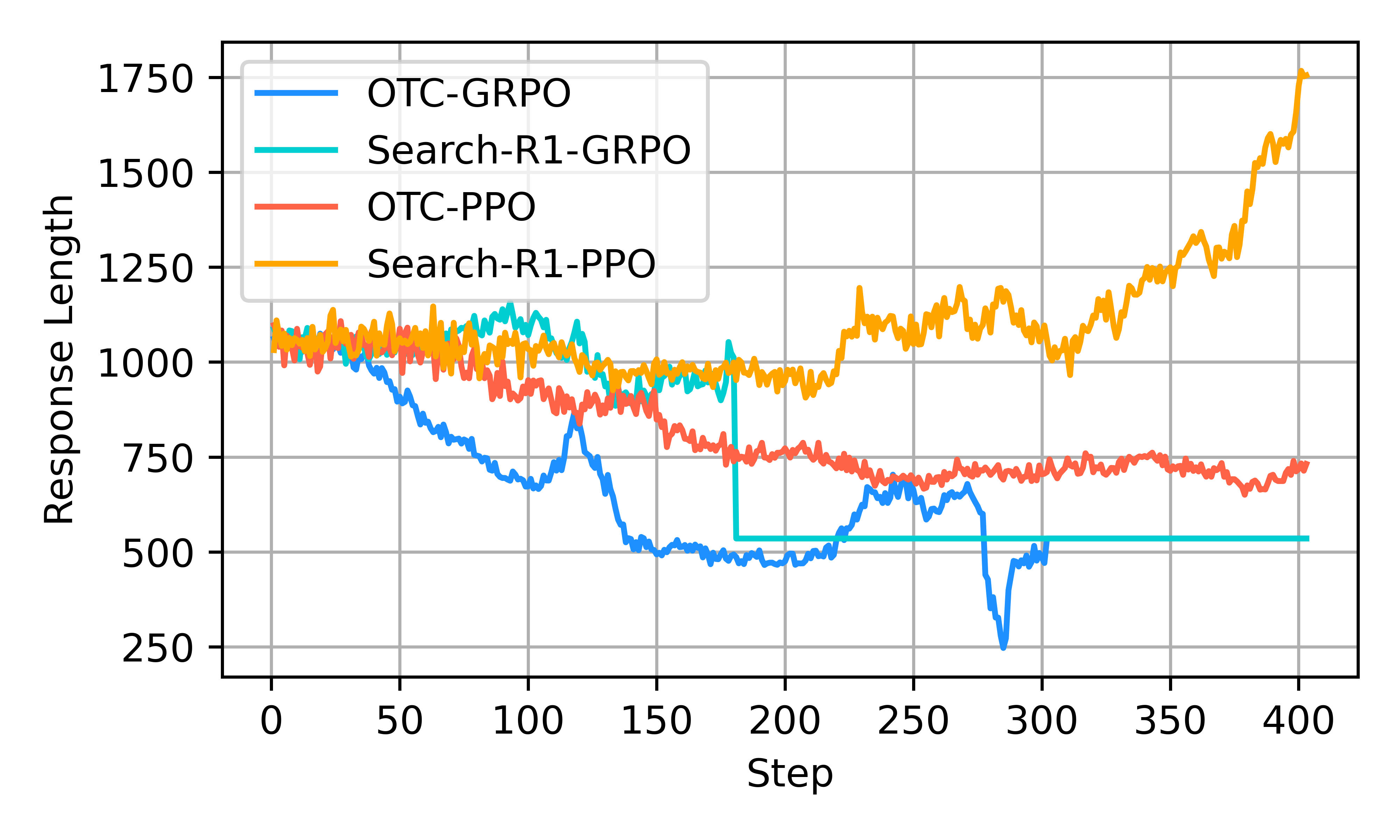}
  \end{subfigure}
  \caption{\textbf{Left}: Changes of number of search calls during the training; \textbf{Middle}: Running time analysis; and \textbf{Right}: Changes of response length during the training.}
  \label{fig:training_analysis}
\end{figure}

\begin{table*}[t]
\centering
\caption{Tool behavioral advantage analysis of OTC-PO against Search-R1. We found that OTC-GRPO and Search-R1-PPO achieves same results with same tool calls in 42.54\% cases of HotpotQA on Qwen2.5-3B-Base model, leading to relatively lower ME.}
\begin{adjustbox}{max width=0.9 \textwidth}
\begin{tabular}{l|ccccc|ccccc}
\toprule
\textbf{Method} & \multicolumn{5}{c}{\textbf{NQ}} & \multicolumn{5}{|c}{\textbf{HotpotQA}} \\  

\cline{2-11} & ME ($\uparrow$) & LE ($\downarrow$) & MA ($\uparrow$) & LA ($\downarrow$) & AE ($\uparrow$) & ME ($\uparrow$) & LE ($\downarrow$) & MA ($\uparrow$) & LA ($\downarrow$) & AE ($\uparrow$) \\

\hline
\multicolumn{3}{l}{\textbf{Qwen2.5-3B(-Base)}} \\

OTC-PPO & 63.55 & 0.02 & 4.60 & 7.67 & 3.74 & 61.78 & 0.53 & 4.30 & 6.34 & 3.35 \\
OTC-GRPO & 65.43 & 2.49 & 6.15 & 3.85 & 5.10 & 35.54 & 5.96 & 11.89 & 4.07 & 3.47 \\


\hline
\multicolumn{3}{l}{\textbf{Qwen2.5-7B(-Base)}} \\

OTC-PPO & 86.2 & 0.03 & 6.45 & 6.92 & 6.45 & 81.49 & 0.08 & 8.60 & 8.68 & 8.53  \\
OTC-GRPO & 82.0 & 0.00 & 7.26 & 10.69 & 7.26 & 79.71 & 0.00 & 8.15 & 11.98 & 8.12 \\

\bottomrule
\end{tabular}
\end{adjustbox}
\label{tab:tool_behavior}
\vspace{-4mm}
\end{table*}

We mainly conduct our analysis using search as a tool in this section and leave more analysis in the Appendix~\ref{appendix:search} and ~\ref{appendix:code}, respectively.

\subsection{Tool Use Behavior Analysis}

\paragraph{Training.} Fig.~\ref{fig:training_analysis} shows the training behaviors of different methods. It is clear that our method not only achieves comparable results with fewer tool calls and shorter responses, but also enables faster and more efficient training optimization. This is particularly important, as it significantly reduces the time and cost associated with real-time tool interactions during training, both in terms of financial expenses and computational overhead. We also find the GRPO is less stable than PPO, as reported by several recent studies~\citep{jin2025searchr1trainingllmsreason, ragen}, and our method can delay the early collapse substantially.

\paragraph{Inference.} We identify several representative tool-use behavior types of our proposed method: i) the answer is same with \textit{less} tool calls compared with baseline (\textit{more efficient}); ii) the answer is same with \textit{more} tool calls compared with baseline (\textit{less efficient}); and iii) our method is able to produce the correct answer whereas the baseline fails to answer correctly (\textit{more effective}); iv) the baseline can produce the correct answer whereas the our method fails to answer correctly (\textit{less effective}); and v) our method is able to produce the correct answer using fewer tool calls whereas the baseline fails to answer correctly (\textit{effective and efficient}).  Table~\ref{tab:tool_behavior} shows the results compared with the baseline Search-R1-PPO \footnote{We use the same baseline to better distinguish OTC-PPO and OTC-GRPO.}. On the efficiency side, it is clear to find that our method achieve same answers with baseline using fewer tool calls in most cases, and more than 80\% when the model becomes larger. On the effectiveness side, we conclude that there is no significant loss when using same RL algorithm, as evidenced by lower gap between MA and LA in OTC-PPO method. Moreover, among all MA cases, it is observed that AE accounts for roughly 80\% in 3B model (i.e., 3.74/4.60~=81\%), and exceeds 95\% on 7B model. These results highlight the great strength of OTC-PO in promoting efficiency while reserving effectiveness.

\subsection{Out-of-domain Evaluation}

\begin{table*}[!t]
\setlength{\belowcaptionskip}{0pt}
    \centering
    \caption{The results of Out-of-Domain (OOD) evaluation of OTC against Search-R1 in EM and TC.}
    \begin{adjustbox}{max width=0.9 \textwidth}
    \begin{tabular}{l| cc|cc|cc|cc|cc}
    \toprule
    \multirow{2}{*}{\textbf{Models}} & \multicolumn{2}{c}{\textbf{TriviaQA}} & \multicolumn{2}{|c}{\textbf{PopQA}} & \multicolumn{2}{|c}{\textbf{2Wiki}} & \multicolumn{2}{|c}{\textbf{Musique}} & \multicolumn{2}{|c}{\textbf{Bamboogle}}  \\  
    \cline{2-11} & EM ($\uparrow$) & TC ($\downarrow$) & EM ($\uparrow$) & TC ($\downarrow$) & EM ($\uparrow$) & TC ($\downarrow$) & EM ($\uparrow$) & TC ($\downarrow$) & EM ($\uparrow$) & TC ($\downarrow$) \\
    \hline
    \multicolumn{3}{l}{\textbf{Qwen2.5-3B(-Base)}} \\

    Search-R1-PPO & 0.566 & 1.580 & 0.425 & 1.631 & 0.258 & 1.675 & 0.051 & 1.922 & 0.063 & 1.766 \\
    Search-R1-GRPO & 0.587 & 1.455 & 0.345 & 1.542 & 0.257 & 1.991 & 0.084 & 2.263 & 0.203 & 1.859 \\

    \hdashline
    OTC-PPO & 0.551 & \textbf{1.008} & 0.409 & \textbf{1.009} & 0.235 & \textbf{1.050} & 0.045 & 1.051 & 0.063 & \textbf{1.016} \\
    OTC-GRPO & \textbf{0.608} & 1.046 & \textbf{0.441} & 1.030 & \textbf{0.341} & 1.561 & \textbf{0.124} & 1.734 & \textbf{0.266} & 1.547 \\

    \hline
    \multicolumn{3}{l}{\textbf{Qwen2.5-7B(-Base)}} \\
    Search-R1-PPO & 0.596 & 3.353 & 0.420 & 3.315 & 0.326 & 4.116 & 0.135 & 4.294 & 0.375 & 3.641 \\
    Search-R1-GRPO & 0.578 & 1.704 & 0.411 & 1.754 & 0.340 & 2.521 & 0.130 & 2.616 & 0.203 & 1.859 \\

    \hdashline
    OTC-PPO & \textbf{0.623} & 1.066 & 0.425 & 1.083 & \textbf{0.363} & 1.868 & \textbf{0.152} & 1.942 & \textbf{0.391} & 1.828 \\
    OTC-GRPO & 0.597 & \textbf{0.430} & \textbf{0.431} & \textbf{0.739} & 0.311 & \textbf{0.938} & 0.130 & \textbf{1.224} & 0.250 & \textbf{0.781} \\
    
    \bottomrule
    \end{tabular}
    \end{adjustbox}
    \label{tab:ood_search_appendix}
    \vspace{-4mm}
\end{table*}

We run out-of-domain evaluation on more QA benchmarks. Fig.~\ref{fig:ood_tp} shows the results of TP and Table~\ref{tab:ood_search_appendix} shows the results of EM and TC. Generally, we find that our method achieves highest TP across all the model and benchmark. More importantly, we observe that OTC-GRPO is better than OTC-PPO on TP and as the model size increases, our method showcases bigger improvement, serving as another evidence of great scalability of our method. Notably, we find that OTC-PPO outperforms the Search-R1-PPO in terms of EM on all benchmarks with Qwen2.5-7B-Base model with much less tool calls, demonstrating the better generalization capabilities of our methods.

\subsection{Case Study}
\label{case_study}

Fig.~\ref{fig:main_case_study} presents a case study to better illustrate the advantages of our proposed OTC-PO. There are several important findings can be drawn from the case study: 1) it is evident that Search-R1 issues multiple redundant and unnecessary search calls. This behavior stems from a lack of incentive to minimize tool usage, as the method is not penalized for excessive tool calls. Consequently, it fails to optimize the efficiency and effectiveness of tool calls, resulting in multiple information-poor queries; 2) both OTC-PPO and OTC-GRPO can reduce the number of unnecessary tool calls significantly but they achieve this in different ways. OTC-PPO utilizes more accurate search by generating information-rich query and OTC-GRPO choose to leverage the internal reasoning capabilities as much as possible if it can be solved without any external tool calls. In summary, there are two valuable and important takeaways: i)\textit{ LLMs tend to overuse external tools and underdevelop their reasoning capabilities when there is no penalty for tool usage - a phenomenon known as cognitive offloading, as exemplified by the behavior of Search-R1;} ii) \textit{Minimizing external tool calls is maximining internal reasoning capabilities of LLMs, leading to more smarter agentic behaviors.} These two findings are crucial and broadly applicable across a wide range of agentic reasoning tasks \footnote{More case studies can be found in Appendix.}.

\section{Conclusion and Future Work}
\label{sec:conclusion}

In this work, we redefine the objective of task for agentic RL not only provide the final correct answer, but also optimize the tool use behavior of model to achieve such goal. We then introduce OTC-PO, a simple yet effective RL framework that explicitly encourages LLMs to generate correct answers with optimal tool calls. Unlike prior work that primarily focuses on final answer correctness, our approach incorporates a tool-integrated reward that accounts for both effectiveness and efficiency of tool usage, thereby promoting tool productivity without sacrificing the accuracy in several benchmarks with different tools. We also find that extensively rely on external tools hinder the development and utilization of  internal reasoning capabilities of the model, and minimizing external tool calls alternatively foster the development of internal reasoning capabilities. Finally, in future work, we aim to extend our framework to more complex agentic tasks involving a broader set of tools and longer-horizon reasoning.




\bibliographystyle{unsrt}
\bibliography{reference.bib}

\begin{thebibliography}{10}

\bibitem{lecun2022path}
Yann LeCun.
\newblock A path towards autonomous machine intelligence version 0.9. 2, 2022-06-27.
\newblock {\em Open Review}, 62(1):1--62, 2022.

\bibitem{openai2024openaio1card}
OpenAI Team.
\newblock Openai o1 system card, 2024.

\bibitem{deepseekai2025deepseekr1}
DeepSeek-AI Team.
\newblock Deepseek-r1: Incentivizing reasoning capability in llms via reinforcement learning, 2025.

\bibitem{qwq32b}
Qwen Team.
\newblock Qwq-32b: Embracing the power of reinforcement learning, March 2025.

\bibitem{wei2023chainofthoughtpromptingelicitsreasoning}
Jason Wei, Xuezhi Wang, Dale Schuurmans, Maarten Bosma, Brian Ichter, Fei Xia, Ed~Chi, Quoc Le, and Denny Zhou.
\newblock Chain-of-thought prompting elicits reasoning in large language models, 2023.

\bibitem{wang2025selfreasoning}
Hongru WANG, Deng Cai, Wanjun Zhong, Shijue Huang, Jeff~Z. Pan, Zeming Liu, and Kam-Fai Wong.
\newblock Self-reasoning language models: Unfold hidden reasoning chains with few reasoning catalyst.
\newblock In {\em Workshop on Reasoning and Planning for Large Language Models}, 2025.

\bibitem{qin2024o1replicationjourneystrategic}
Yiwei Qin, Xuefeng Li, Haoyang Zou, Yixiu Liu, Shijie Xia, Zhen Huang, Yixin Ye, Weizhe Yuan, Hector Liu, Yuanzhi Li, and Pengfei Liu.
\newblock O1 replication journey: A strategic progress report -- part 1, 2024.

\bibitem{zeng2025simplerlzooinvestigatingtamingzero}
Weihao Zeng, Yuzhen Huang, Qian Liu, Wei Liu, Keqing He, Zejun Ma, and Junxian He.
\newblock Simplerl-zoo: Investigating and taming zero reinforcement learning for open base models in the wild, 2025.

\bibitem{gou2024tora}
Zhibin Gou, Zhihong Shao, Yeyun Gong, yelong shen, Yujiu Yang, Minlie Huang, Nan Duan, and Weizhu Chen.
\newblock To{RA}: A tool-integrated reasoning agent for mathematical problem solving.
\newblock In {\em The Twelfth International Conference on Learning Representations}, 2024.

\bibitem{li2025startselftaughtreasonertools}
Chengpeng Li, Mingfeng Xue, Zhenru Zhang, Jiaxi Yang, Beichen Zhang, Xiang Wang, Bowen Yu, Binyuan Hui, Junyang Lin, and Dayiheng Liu.
\newblock Start: Self-taught reasoner with tools, 2025.

\bibitem{wang-etal-2023-large}
Hongru Wang, Minda Hu, Yang Deng, Rui Wang, Fei Mi, Weichao Wang, Yasheng Wang, Wai-Chung Kwan, Irwin King, and Kam-Fai Wong.
\newblock Large language models as source planner for personalized knowledge-grounded dialogues.
\newblock In Houda Bouamor, Juan Pino, and Kalika Bali, editors, {\em Findings of the Association for Computational Linguistics: EMNLP 2023}, pages 9556--9569, Singapore, December 2023. Association for Computational Linguistics.

\bibitem{qian2025smartselfawareagenttool}
Cheng Qian, Emre~Can Acikgoz, Hongru Wang, Xiusi Chen, Avirup Sil, Dilek Hakkani-T{\"u}r, Gokhan Tur, and Heng Ji.
\newblock Smart: Self-aware agent for tool overuse mitigation.
\newblock {\em arXiv preprint arXiv:2502.11435}, 2025.

\bibitem{tool_learning_tut}
Hongru Wang, Yujia Qin, Yankai Lin, Jeff~Z. Pan, and Kam-Fai Wong.
\newblock Empowering large language models: Tool learning for real-world interaction.
\newblock In {\em Proceedings of the 47th International ACM SIGIR Conference on Research and Development in Information Retrieval}, SIGIR '24, page 2983–2986, New York, NY, USA, 2024. Association for Computing Machinery.

\bibitem{jin2025searchr1trainingllmsreason}
Bowen Jin, Hansi Zeng, Zhenrui Yue, Dong Wang, Hamed Zamani, and Jiawei Han.
\newblock Search-r1: Training llms to reason and leverage search engines with reinforcement learning, 2025.

\bibitem{li2025torlscalingtoolintegratedrl}
Xuefeng Li, Haoyang Zou, and Pengfei Liu.
\newblock Torl: Scaling tool-integrated rl, 2025.

\bibitem{chen2025researchlearningreasonsearch}
Mingyang Chen, Tianpeng Li, Haoze Sun, Yijie Zhou, Chenzheng Zhu, Haofen Wang, Jeff~Z. Pan, Wen Zhang, Huajun Chen, Fan Yang, Zenan Zhou, and Weipeng Chen.
\newblock Research: Learning to reason with search for llms via reinforcement learning, 2025.

\bibitem{risko2016cognitive}
Evan~F Risko and Sam~J Gilbert.
\newblock Cognitive offloading.
\newblock {\em Trends in cognitive sciences}, 20(9):676--688, 2016.

\bibitem{qwen2025qwen25technicalreport}
Qwen, :, An~Yang, Baosong Yang, Beichen Zhang, Binyuan Hui, Bo~Zheng, Bowen Yu, Chengyuan Li, Dayiheng Liu, Fei Huang, Haoran Wei, Huan Lin, Jian Yang, Jianhong Tu, Jianwei Zhang, Jianxin Yang, Jiaxi Yang, Jingren Zhou, Junyang Lin, Kai Dang, Keming Lu, Keqin Bao, Kexin Yang, Le~Yu, Mei Li, Mingfeng Xue, Pei Zhang, Qin Zhu, Rui Men, Runji Lin, Tianhao Li, Tianyi Tang, Tingyu Xia, Xingzhang Ren, Xuancheng Ren, Yang Fan, Yang Su, Yichang Zhang, Yu~Wan, Yuqiong Liu, Zeyu Cui, Zhenru Zhang, and Zihan Qiu.
\newblock Qwen2.5 technical report, 2025.

\bibitem{schulman2017proximal}
John Schulman, Filip Wolski, Prafulla Dhariwal, Alec Radford, and Oleg Klimov.
\newblock Proximal policy optimization algorithms.
\newblock {\em arXiv preprint arXiv:1707.06347}, 2017.

\bibitem{shao2024deepseekmath}
Zhihong Shao, Peiyi Wang, Qihao Zhu, Runxin Xu, Junxiao Song, Xiao Bi, Haowei Zhang, Mingchuan Zhang, YK~Li, Y~Wu, et~al.
\newblock Deepseekmath: Pushing the limits of mathematical reasoning in open language models.
\newblock {\em arXiv preprint arXiv:2402.03300}, 2024.

\bibitem{qian-etal-2023-creator}
Cheng Qian, Chi Han, Yi~Fung, Yujia Qin, Zhiyuan Liu, and Heng Ji.
\newblock {CREATOR}: Tool creation for disentangling abstract and concrete reasoning of large language models.
\newblock In Houda Bouamor, Juan Pino, and Kalika Bali, editors, {\em Findings of the Association for Computational Linguistics: EMNLP 2023}, pages 6922--6939, Singapore, December 2023. Association for Computational Linguistics.

\bibitem{wang-etal-2024-appbench}
Hongru Wang, Rui Wang, Boyang Xue, Heming Xia, Jingtao Cao, Zeming Liu, Jeff~Z. Pan, and Kam-Fai Wong.
\newblock {A}pp{B}ench: Planning of multiple {API}s from various {APP}s for complex user instruction.
\newblock In Yaser Al-Onaizan, Mohit Bansal, and Yun-Nung Chen, editors, {\em Proceedings of the 2024 Conference on Empirical Methods in Natural Language Processing}, pages 15322--15336, Miami, Florida, USA, November 2024. Association for Computational Linguistics.

\bibitem{qian2024investigate}
Cheng Qian, Shihao Liang, Yujia Qin, Yining Ye, Xin Cong, Yankai Lin, Yesai Wu, Zhiyuan Liu, and Maosong Sun.
\newblock Investigate-consolidate-exploit: A general strategy for inter-task agent self-evolution.
\newblock {\em arXiv preprint arXiv:2401.13996}, 2024.

\bibitem{yuan2024craftcustomizingllmscreating}
Lifan Yuan, Yangyi Chen, Xingyao Wang, Yi~R. Fung, Hao Peng, and Heng Ji.
\newblock Craft: Customizing llms by creating and retrieving from specialized toolsets, 2024.

\bibitem{qian2024escapebench}
Cheng Qian, Peixuan Han, Qinyu Luo, Bingxiang He, Xiusi Chen, Yuji Zhang, Hongyi Du, Jiarui Yao, Xiaocheng Yang, Denghui Zhang, et~al.
\newblock Escapebench: Pushing language models to think outside the box.
\newblock {\em arXiv preprint arXiv:2412.13549}, 2024.

\bibitem{wang2025selfdcreasonactself}
Hongru Wang, Boyang Xue, Baohang Zhou, Tianhua Zhang, Cunxiang Wang, Huimin Wang, Guanhua Chen, and Kam fai Wong.
\newblock Self-dc: When to reason and when to act? self divide-and-conquer for compositional unknown questions, 2025.

\bibitem{qin2023toolllmfacilitatinglargelanguage}
Yujia Qin, Shihao Liang, Yining Ye, Kunlun Zhu, Lan Yan, Yaxi Lu, Yankai Lin, Xin Cong, Xiangru Tang, Bill Qian, Sihan Zhao, Lauren Hong, Runchu Tian, Ruobing Xie, Jie Zhou, Mark Gerstein, Dahai Li, Zhiyuan Liu, and Maosong Sun.
\newblock Toolllm: Facilitating large language models to master 16000+ real-world apis, 2023.

\bibitem{qian2024toolink}
Cheng Qian, Chenyan Xiong, Zhenghao Liu, and Zhiyuan Liu.
\newblock Toolink: Linking toolkit creation and using through chain-of-solving on open-source model.
\newblock In {\em Proceedings of the 2024 Conference of the North American Chapter of the Association for Computational Linguistics: Human Language Technologies (Volume 1: Long Papers)}, pages 831--854, 2024.

\bibitem{chen2025advancingtoolaugmentedlargelanguage}
Sijia Chen, Yibo Wang, Yi-Feng Wu, Qing-Guo Chen, Zhao Xu, Weihua Luo, Kaifu Zhang, and Lijun Zhang.
\newblock Advancing tool-augmented large language models: Integrating insights from errors in inference trees, 2025.

\bibitem{feng2025retoolreinforcementlearningstrategic}
Jiazhan Feng, Shijue Huang, Xingwei Qu, Ge~Zhang, Yujia Qin, Baoquan Zhong, Chengquan Jiang, Jinxin Chi, and Wanjun Zhong.
\newblock Retool: Reinforcement learning for strategic tool use in llms, 2025.

\bibitem{qian2025toolrlrewardisalltool}
Cheng Qian, Emre~Can Acikgoz, Qi~He, Hongru Wang, Xiusi Chen, Dilek Hakkani-T{\"u}r, Gokhan Tur, and Heng Ji.
\newblock Toolrl: Reward is all tool learning needs.
\newblock {\em arXiv preprint}, 2025.

\bibitem{shen-etal-2024-smartcal}
Yuanhao Shen, Xiaodan Zhu, and Lei Chen.
\newblock {SMARTCAL}: An approach to self-aware tool-use evaluation and calibration.
\newblock In Franck Dernoncourt, Daniel Preo{\c{t}}iuc-Pietro, and Anastasia Shimorina, editors, {\em Proceedings of the 2024 Conference on Empirical Methods in Natural Language Processing: Industry Track}, pages 774--789, Miami, Florida, US, November 2024. Association for Computational Linguistics.

\bibitem{li2025adaptivetooluselarge}
Wenjun Li, Dexun Li, Kuicai Dong, Cong Zhang, Hao Zhang, Weiwen Liu, Yasheng Wang, Ruiming Tang, and Yong Liu.
\newblock Adaptive tool use in large language models with meta-cognition trigger, 2025.

\bibitem{arora2025traininglanguagemodelsreason}
Daman Arora and Andrea Zanette.
\newblock Training language models to reason efficiently, 2025.

\bibitem{ragen}
Zihan Wang, Kangrui Wang, Qineng Wang, Pingyue Zhang, Linjie Li, Zhengyuan Yang, Kefan Yu, Minh~Nhat Nguyen, Licheng Liu, Eli Gottlieb, Monica Lam, Yiping Lu, Kyunghyun Cho, Jiajun Wu, Li~Fei-Fei, Lijuan Wang, Yejin Choi, and Manling Li.
\newblock Ragen: Understanding self-evolution in llm agents via multi-turn reinforcement learning, 2025.

\bibitem{laud2004theory}
Adam~Daniel Laud.
\newblock {\em Theory and application of reward shaping in reinforcement learning}.
\newblock University of Illinois at Urbana-Champaign, 2004.

\bibitem{rl_bert}
Hongru Wang, Huimin Wang, Zezhong Wang, and Kam-Fai Wong.
\newblock Integrating pretrained language model for dialogue policy evaluation.
\newblock In {\em ICASSP 2022 - 2022 IEEE International Conference on Acoustics, Speech and Signal Processing (ICASSP)}, pages 6692--6696, 2022.

\bibitem{aggarwal2025l1controllinglongreasoning}
Pranjal Aggarwal and Sean Welleck.
\newblock L1: Controlling how long a reasoning model thinks with reinforcement learning, 2025.

\bibitem{wang2025harnessingreasoningeconomysurvey}
Rui Wang, Hongru Wang, Boyang Xue, Jianhui Pang, Shudong Liu, Yi~Chen, Jiahao Qiu, Derek~Fai Wong, Heng Ji, and Kam-Fai Wong.
\newblock Harnessing the reasoning economy: A survey of efficient reasoning for large language models, 2025.

\bibitem{hou2025thinkprunepruninglongchainofthought}
Bairu Hou, Yang Zhang, Jiabao Ji, Yujian Liu, Kaizhi Qian, Jacob Andreas, and Shiyu Chang.
\newblock Thinkprune: Pruning long chain-of-thought of llms via reinforcement learning, 2025.

\bibitem{ouyang2022traininglanguagemodelsfollow}
Long Ouyang, Jeff Wu, Xu~Jiang, Diogo Almeida, Carroll~L. Wainwright, Pamela Mishkin, Chong Zhang, Sandhini Agarwal, Katarina Slama, Alex Ray, John Schulman, Jacob Hilton, Fraser Kelton, Luke Miller, Maddie Simens, Amanda Askell, Peter Welinder, Paul Christiano, Jan Leike, and Ryan Lowe.
\newblock Training language models to follow instructions with human feedback, 2022.

\end{thebibliography}


\appendix
\clearpage

\section{Related Work}

\paragraph{Reward Shaping in RL.}  Reward shaping plays a critical role in reinforcement learning, as it directly shapes the behavior the model learns to optimize~\citep{laud2004theory, rl_bert}. Recent advances have introduced several reward signals for LLMs to consider the correctness and the internal knowledge-only reasoning efficiency together such as the length of responses and difficulty of the questions \citep{aggarwal2025l1controllinglongreasoning, arora2025traininglanguagemodelsreason, wang2025harnessingreasoningeconomysurvey}. For instances, Length Controlled Policy Optimization (LCPO)~\citep{aggarwal2025l1controllinglongreasoning} is proposed to satisfy the length constraints while optimizing reasoning performance and some efforts try to dynamically allocate inference time compute based on task complexity~\citep{arora2025traininglanguagemodelsreason, hou2025thinkprunepruninglongchainofthought}. 

\section{Tool-integrated Reinforcement Learning}
\label{appendix_trl}

\paragraph{PPO in TRL.} Proximal Policy Optimization (PPO) is a widely used policy-gradient method in lots of tasks~\citep{ouyang2022traininglanguagemodelsfollow}. Given our formulation, PPO updates the policy, using trajectories sampled from the previous policy, and maximizes the following objective:

\begin{equation}
\small
\begin{aligned}
\mathcal{J}_{\text{PPO}}(\theta) = \mathbb{E}_{q \sim \mathcal{D},\, y \sim \pi_{\text{old}}} \Bigg[
\frac{1}{\sum_{t=1}^{|y|} \mathbb{I}(y_t)} \sum_{t=1}^{|y|} \mathbb{I}(y_t) \cdot 
\min\Bigg(&\rho_t A_t,\,
\text{clip}(\rho_t, 1 - \epsilon, 1 + \epsilon) A_t
\Bigg) \Bigg],
\end{aligned}
\end{equation}

where $\pi_{\theta}$ and $\pi_{old}$ are current and previous policy models, and $p_t$=$\frac{\pi_{\theta} (y_t | q,y_{<t}; \mathcal{E})}{\pi_{old} (y_t | q,y_{<t}; \mathcal{E})}$. Here, $\mathbb{I} (y_t)$ is an indicator function marking whether token $y_t$ is generated by the model (i.e., $r_i$ and $tc_i$) or returned from the environment $\mathcal{E}$ (i.e., $o_i$).  The advantage estimate $A_t$ is computed via Generalized Advantage Estimation (GAE) and $\epsilon$ a PPO clipping threshold to constrain the policy update.

\paragraph{GRPO in TRL.} To improve the stability of policy optimization in language models and avoid reliance on an additional value function approximation, Group Relative Policy Optimization (GRPO) is introduced which uses the relative ranking of multiple sampled outputs as a baseline for computing advantages, rather than fitting a value function. For each input question $q$, GRPO samples a group of $G$ response $\{y_1, y_2, ..., y_G\}$ from the reference policy $\pi_{ref}$. The trained policy $\pi_\theta$ is then updated by maximizing the following objective:

\begin{equation}
\begin{aligned}
    \mathcal{J}_{\text{GRPO}}(\theta) = 
    \mathbb{E}_{q \sim \mathcal{D},\, \{y_i\}_{i=1}^G \sim \tau_{\text{old}}(\cdot \mid q; \mathcal{E})}
    \bigg[ \frac{1}{G} \sum_{i=1}^G \frac{1}{\sum_{t=1}^{|y_i|} \mathbb{I}(y_{i,t})} 
    \sum_{t=1}^{|y_i|} \mathbb{I}(y_{i,t}) \\
    \cdot \min\left(
        p_t \hat{A}_{i,t},\
        \text{clip}\left(
            p_t,
            1 - \epsilon,\ 1 + \epsilon
        \right) \hat{A}_{i,t}
    \right)
    \bigg] 
    - \beta\, \mathbb{D}_{\text{KL}}[\pi_\theta \,\|\, \pi_{\text{ref}}],
\end{aligned}
\end{equation}

Here, $p_t$ = $\frac{\pi_\theta(y_{i,t} \mid x, y_{i,<t}; \mathcal{R})}{\pi_{\text{old}}(y_{i,t} \mid x, y_{i,<t}; \mathcal{R})}$, $\hat{A}_{i,t}$ denotes the advantage at token $t$ in response $y_i$, computed based on the relative ranking of rewards within the group, $\beta$ controlling the KL-regularization strength. The clipping threshold \( \epsilon \) ensures stable updates.

\section{Search Template}

\begin{tcolorbox}
Answer the given question. You must conduct reasoning inside <think> and </think>
first every time you get new information. After reasoning, if you find you lack some
knowledge, you can call a search engine tool by <search> query </search>, and it will
return the top searched results between <observation> and </observation>. \textcolor{red}{You need to make every search call count and gain helpful results.} If you find no further external knowledge
needed, you can directly provide the answer inside <answer> and </answer> without
detailed illustrations. For example, <answer> xxx </answer>. Question: question.
\end{tcolorbox}

\section{Reward Function}
\label{appendix:reward}

\subsection{Figure Illustration}

\begin{figure}[htbp]
  \centering
  \begin{subfigure}[b]{0.2\textwidth}
    \includegraphics[height=3cm, keepaspectratio]{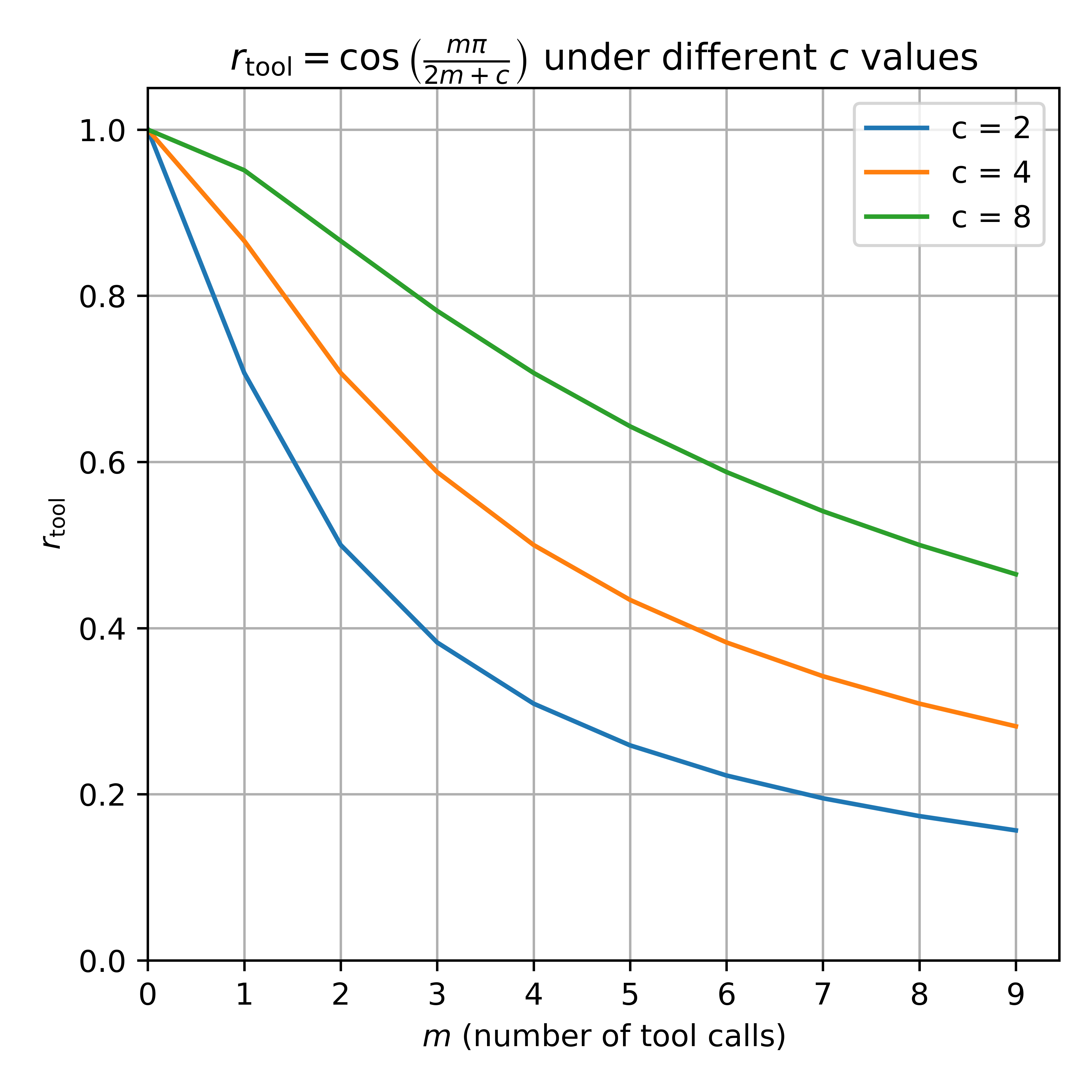}
  \end{subfigure}
  \begin{subfigure}[b]{0.7\textwidth}
    \includegraphics[height=3cm, keepaspectratio]{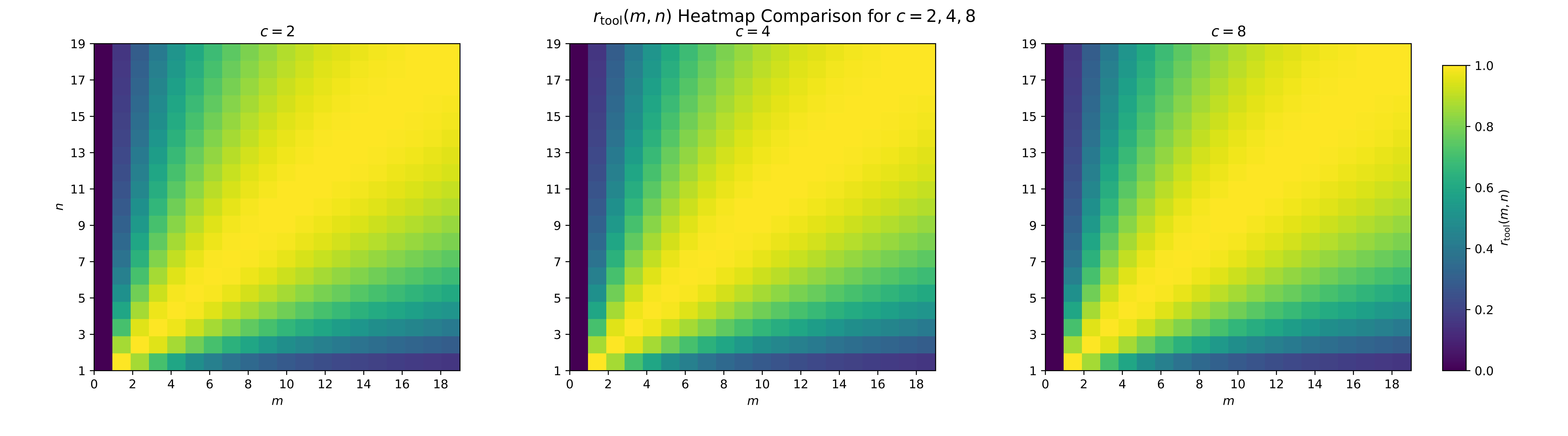}
  \end{subfigure}
  \caption{\textbf{Left}: the illustration of tool-use reward functions in OTC-PPO; and \textbf{Right}: the illustration of tool-use reward functions in OTC-GRPO.}
  \label{fig:reward_illustrations}
\end{figure}

We draw the two types of $r_{tool}$ defined in the main content for better understanding. Fig.~\ref{fig:reward_illustrations} shows the illustration of these reward functions for OTC-PPO and OTC-GRPO respectively. It is very clear that: 1) Left: as the number of tool calls increases, the $r_{tool}$ decreases accordingly. Thus when multiple trajectories leads to correct answer, the one that use less tools will get higher reward and the one that do not use tool will get the highest reward; 2) Right: It is obvious that the diagonal achieves the highest reward as the number of tool calls is the optimal one, and as the $m$ increases, the color brightness becomes larger since multiple tool calls are involved. 

\subsection{Theoretical Justification}
\label{sec:theoretical_proof}

We mainly follow the justification (Section 4.2) in the paper ~\citep{arora2025traininglanguagemodelsreason} to showcase our proposed reward design leads to reducing the tool calls without compromising accuracy. We strongly encourage readers to refer to the previous paper for complete details. All assumptions made in that work still hold in our setting, with the only difference being that our response is defined as $y$ = $(\tau, a)$, which includes both the tool-integrated reasoning trajectory and the final answer, rather than language-only reasoning.

Let $\theta^*_{eff}$ denote the population-level parameters of the policy models obtained by maximizing Equation~\ref{eq:tool_total_reward}, i.e.,

\begin{equation}
\label{eq:eff_objective}
\theta^{\star}_{\text{eff}} = \arg\max_{\theta} \left\{ 
\mathbb{E}_{x \sim p} \mathbb{E}_{y \sim p_{\theta}(x)} \left[ 
(\alpha * r_{tool}) 
\right] \right\}
\end{equation}

as $r_\phi (q, y)$ is mainly the indicator function about the correctness and format. Therefore, the population-level maximizer $p_{\theta^*_{eff}}$ is as accurate as the population-level maximizer $p^*_{\theta}$ and Acc ($p^*_{\theta_{eff}}$) = 1.


\subsection{Generalization of Reward}

We use two major forms of $r_\phi$: i) $r^1_{\phi} = r_{correct}$; and ii) $r^2_{\phi} = r_{correct} + r_{format}$ to illustrate the generalization of our proposed tool-integrated reward design $r^{tool}_\phi = \alpha * r_{tool} * r_\phi$. 

\textit{If the answer is correct}, the $r^1_\phi$ and $r^2_\phi$ are both positive and fixed for all questions, therefore the $r^{tool}_\phi$ is only determined by the number of $\alpha$ and $r_{tool}$. Thus it can assign higher score for less tools and lower score for more tools by definition of $r_{tool}$. 

\textit{If the answer is wrong}, the $r^1_\phi$ will be 0, and therefore disables the influence of $r_{tool}$, reducing the reward hacking issue. The $r^2_\phi$ will be $r_{format}$ and then $r^{tool}_\phi$ becomes $\alpha * r_{tool} * r_{format}$ which is acceptable. Considering two cases of $r_{format}$, if $r_{format}$ is 0, then the final reward is also 0; and if $r_{format}$ is a positive, the $r^{tool}_\phi$ still holds as less tools will be assigned more rewards. This is reasonable since less tool calls means less cost especially when the answer is wrong. We note that the reward gap here between wrong answer (i.e., $r_{format}$) and correct answer $(i.e., r_{correct} + r_{format})$ is significant for the policy model to learn the desirable behavior.


\section{Analysis of Search as Tool}
\label{appendix:search}




\begin{figure}
\centering
\begin{subfigure}[b]{1.0\textwidth}
    \includegraphics[width=1.0\textwidth, height=3cm, keepaspectratio]{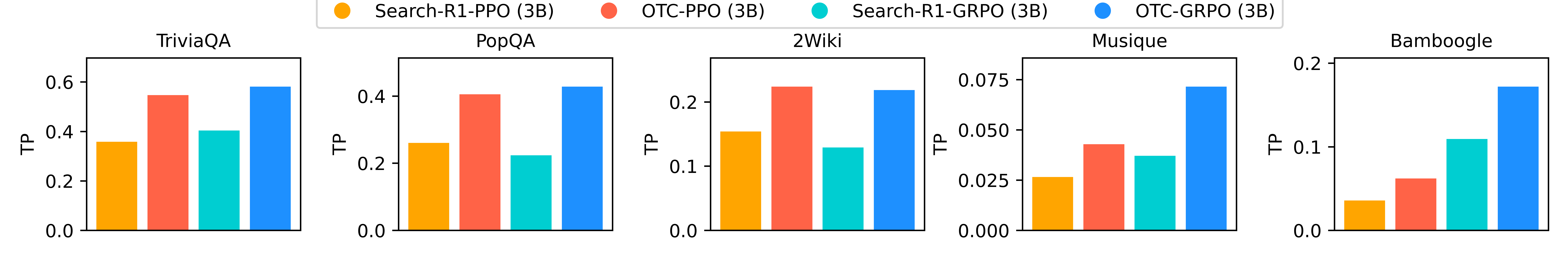}
    \caption{Qwen2.5-3B-Base}
  \end{subfigure}

  \vspace{0.8em} 
  
  \begin{subfigure}[b]{1.0\textwidth}
    \includegraphics[width=1.0\textwidth, height=3cm, keepaspectratio]{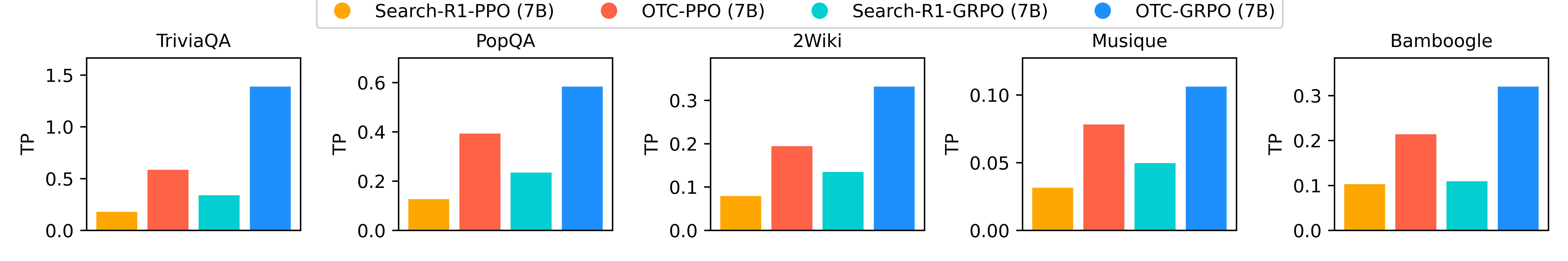}
    \caption{Qwen2.5-7B-Base}
  \end{subfigure}
\caption{The Out-of-domain performance of OTC-PO and Search-R1 in TP.}
\label{fig:ood_tp}
\end{figure}

\subsection{Base v.s. Instruct Model}

\begin{table*}[!t]
\setlength{\belowcaptionskip}{0pt}
    \centering
    \caption{The results of OTC-PO under Qwen2.5-7B-Base and Qwen2.5-7B-Instruct models.}
    \setlength{\tabcolsep}{10pt} 
    \begin{adjustbox}{max width=0.8 \textwidth}
    \begin{tabular}{l| ccc|ccc}
    \toprule
    \multirow{2}{*}{\textbf{Models}} & \multicolumn{3}{c}{\textbf{NQ}} & \multicolumn{3}{|c}{\textbf{HotpotQA}} \\  
    \cline{2-7} & EM ($\uparrow$) & TC ($\downarrow$) & TP ($\uparrow$) & EM ($\uparrow$) & TC ($\downarrow$) & TP ($\uparrow$) \\
    \hline
    \multicolumn{3}{l}{\textbf{Qwen2.5-7B(-Base)}} \\

    OTC-PPO & \textbf{0.446} & 1.040 & 0.429 & 0.383 & 1.464 & 0.262 \\
    OTC-GRPO & 0.444 & \textbf{0.990} & \textbf{0.448} & 0.366 & \textbf{1.005} & \textbf{0.364} \\

    \multicolumn{3}{l}{\textbf{Qwen2.5-7B(-Instruct)}} \\
    OTC-PPO & 0.389 & 1.404 & 0.277 & 0.381 & 1.880 & 0.203 \\
    OTC-GRPO & 0.429 & 1.322 & 0.325 & \textbf{0.386} & 1.956 & 0.197 \\
    
    \bottomrule
    \end{tabular}
    \end{adjustbox}
    \label{tab:base_instruct}
\end{table*}

We additionally compare the performance of OTC-PO using Qwen-7B-Base and Qwen-7B-Instruct models. Table~\ref{tab:base_instruct} shows the final results. It is observed that i) Base models mostly lead to better performance, which is also consistent with results in the Search-R1 paper; ii) Instruct models tends to use more tools compared with base model while the TC still is much lower than the Search-R1. We attribute this to the Instruct model tends to memorize specific pattern during supervised fine-tuning and easily to be triggered to use tools even when it does not need it.

\subsection{The Effects of $\alpha$}

\begin{table*}[!t]
\setlength{\belowcaptionskip}{0pt}
    \centering
    \caption{The results of OTC-PO with different $\alpha$ on Qwen2.5-7B-Base model}
    \begin{adjustbox}{max width=0.9 \textwidth}
    \begin{tabular}{l| ccc|ccc}
    \toprule
    \multirow{2}{*}{\textbf{Models}} & \multicolumn{3}{c}{\textbf{NQ}} & \multicolumn{3}{|c}{\textbf{HotpotQA}} \\  
    \cline{2-7} & EM ($\uparrow$) & TC ($\downarrow$) & TP ($\uparrow$) & EM ($\uparrow$) & TC ($\downarrow$) & TP ($\uparrow$) \\
    \hline
    $\alpha$ = 1 & 0.446 & 1.040 & 0.429 & 0.383 & 1.464  & 0.262 \\
    $\alpha$ = 2 & 0.354 & 1.571 & 0.225 & 0.320 & 2.062 & 0.155 \\
    $\alpha$ = 3 & 0.389 & 1.530 & 0.254 & 0.340 & 1.948 & 0.175 \\
    
    \bottomrule
    \end{tabular}
    \end{adjustbox}
    \label{tab:alpha_effects}
\end{table*}

Table~\ref{tab:alpha_effects} shows the performance of OTC-PO with different $\alpha$. It can be found that increasing $\alpha$ can not leads to better results. We further check the reward changes during the training and find that it becomes harder for the model to learn the desired behavior when using a larger $\alpha$, because a single tool call can cause large fluctuations in the reward, especially considering the relatively lower $C$. We suspect the $\alpha$ will be more important in much longer tool-integrated reasoning trajectories.

\section{Analysis of Code as Tool}
\label{appendix:code}

\subsection{The Effects of $C$}

\begin{table*}[!t]
\setlength{\belowcaptionskip}{0pt}
    \centering
    \setlength{\tabcolsep}{5pt} 
    \caption{The effects of $C$ on Qwen2.5-Math-7B Base model.}
    \begin{adjustbox}{max width=0.9 \textwidth}
    \begin{tabular}{l| ccc|ccc}
    \toprule
    \multirow{2}{*}{\textbf{Models}} & \multicolumn{3}{c}{\textbf{AIME24}} & \multicolumn{3}{|c}{\textbf{AIME25}} \\  
    \cline{2-7} & EM ($\uparrow$) & TC ($\downarrow$) & TP ($\uparrow$) & EM ($\uparrow$) & TC ($\downarrow$) & TP ($\uparrow$) \\
    \hline
    \multicolumn{3}{l}{\textbf{\textit{C = 1}}} \\
    ToRL-GRPO & 30.0 & 0.9 & 33.3 & 26.7 & 0.9 & 29.7 \\
    OTC-GRPO & 35.8 & 0.8 & 44.8 (\textcolor{red!60}{$\blacktriangle$ 34.5\%}) & 26.7 & 0.8 & 33.4 (\textcolor{red!60}{$\blacktriangle$ 11.1\%}) \\
    
    \hdashline
    \multicolumn{3}{l}{\textbf{\textit{C = 2}}} \\
    ToRL-GRPO & 33.3 & 1.4 & 23.8 & 25.8 & 1.4 & 18.4 \\
    OTC-GRPO & 33.3 & 1.0 & 33.3 (\textcolor{red!60}{$\blacktriangle$ 39.9\%}) & 23.3 & 0.7 & 33.3 (\textcolor{red!60}{$\blacktriangle$ 81.0\%}) \\

    \hdashline
    \multicolumn{3}{l}{\textbf{\textit{C = 3}}} \\
    ToRL-GRPO & 36.7 & 2.1 & 17.5 & 26.7 & 2.1 & 12.7 \\
    OTC-GRPO & 36.7 & 0.7 & 52.4 (\textcolor{red!60}{$\blacktriangle$ 199.4\%}) & 23.3 & 0.8 & 29.1 (\textcolor{red!60}{$\blacktriangle$ 129.1\%}) \\
    
    \bottomrule
    \end{tabular}
    \end{adjustbox}
    \label{tab:c_effect_code}
\end{table*}


\begin{figure}[t]
  \centering
  \begin{subfigure}[b]{0.45\textwidth}
    \includegraphics[width=\linewidth]{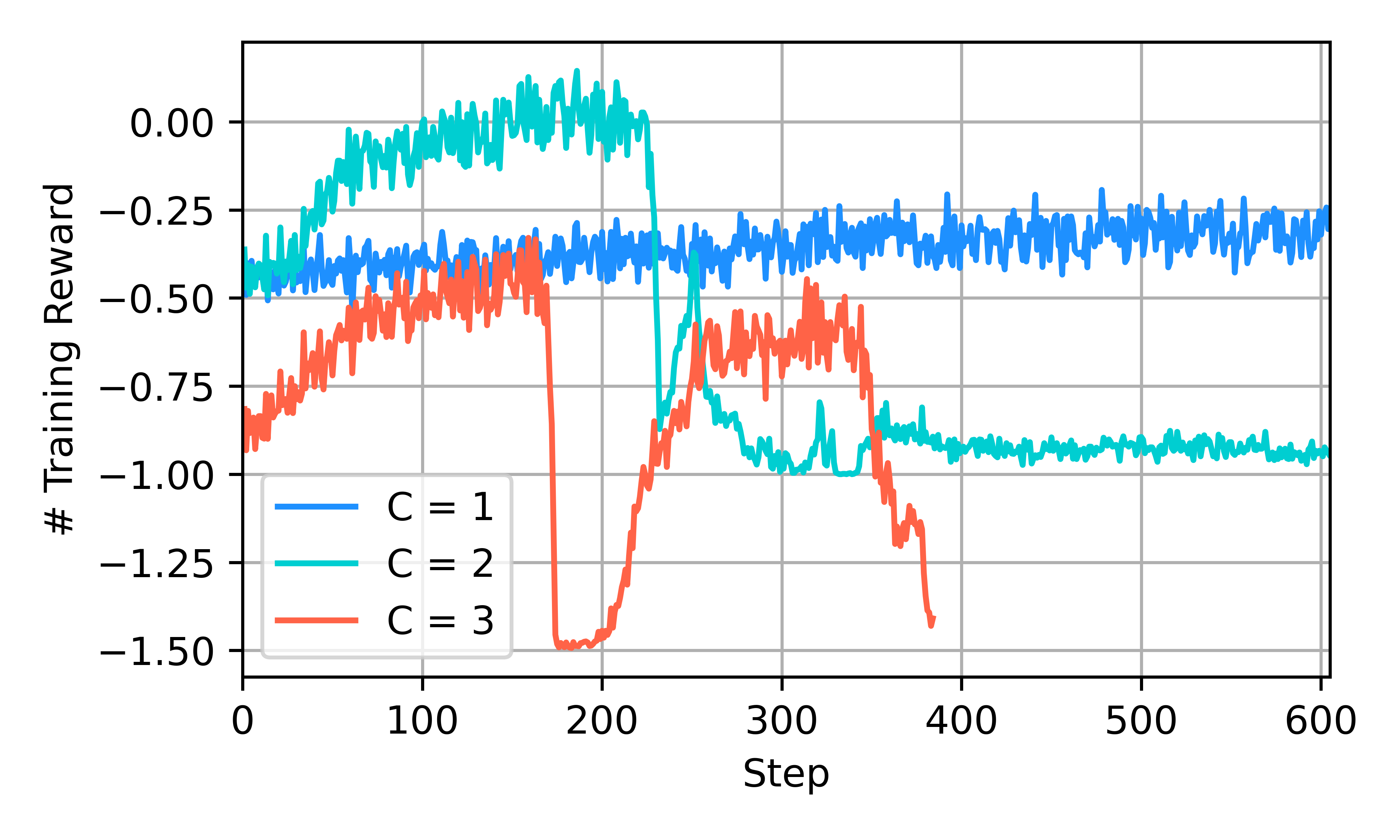}
  \end{subfigure}
  \vspace{5mm}
  \begin{subfigure}[b]{0.45\textwidth}
    \includegraphics[width=\linewidth]{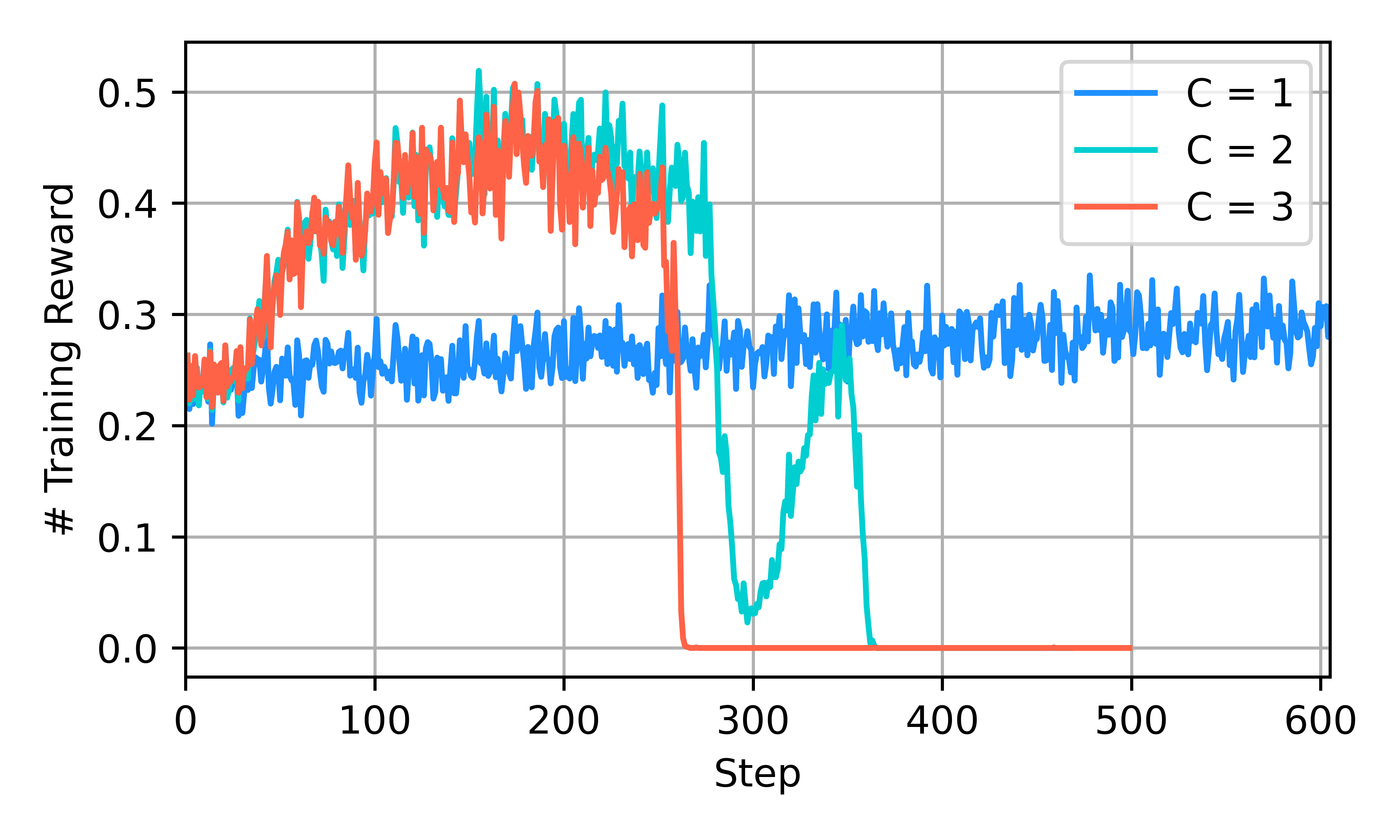}
  \end{subfigure}
  \caption{\textbf{Left}: Training reward changes of ToRL-GRPO; and \textbf{Right}: Training reward changes of OTC-GRPO.}
  \label{fig:c_effects_reward}
\end{figure}

Table~\ref{tab:c_effect_code} shows the effects of $C$ in ToRL. It is observed that i) the larger $C$ always leads to more tool calls on ToRL-GRPO if we only consider the final correctness as the reward and do not penalize the tool use behaviors of LLMs; ii) OTC-GRPO achieves more stable tool calls which is more reasonable as the optimal number of tool calls should not be affected by $C$, and leads to bigger TP improvement as $C$ increases. Further, Fig.~\ref{fig:c_effects_reward} shows the rewards changes with different $C$. It can be found that GRPO tends to 
collapse earlier with bigger $C$ and our OTC-PO can delay this, as also observed using search as tool.

\subsection{Additional Results}

\begin{table*}[!t]
\setlength{\belowcaptionskip}{0pt}
    \centering
    \caption{The results of OTC-PO on MATH when $C$ = 3.}
    \setlength{\tabcolsep}{5pt} 
    \begin{adjustbox}{max width=0.9 \textwidth}
    \begin{tabular}{l| ccc}
    \toprule
    \multirow{2}{*}{\textbf{Models}} & \multicolumn{3}{c}{\textbf{MATH}} \\  
    \cline{2-4} & EM ($\uparrow$) & TC ($\downarrow$) & TP ($\uparrow$)  \\
    \hline
    \multicolumn{3}{l}{\textbf{Qwen2.5-Math-1.5B(-Base)}} \\
    ToRL-GRPO & 52.2 & 1.8 & 29.0 \\
    OTC-GRPO & 53.7 & 1.1 (\textcolor{green!60}{$\blacktriangledown$ 38.9\%}) & 48.8 (\textcolor{red!60}{$\blacktriangle$ 68.28\%}) \\

    \hdashline
    \multicolumn{3}{l}{\textbf{Qwen2.5-Math-7B(-Base)}} \\
    ToRL-GRPO & 59.7 & 1.49 & 40.07 \\
    OTC-GRPO & 61.9 & 0.68 (\textcolor{green!60}{$\blacktriangledown$ 54.4\%}) & 91.03 (\textcolor{red!60}{$\blacktriangle$ 127.2\%}) \\

    \bottomrule
    \end{tabular}
    \end{adjustbox}
    \label{tab:math_results}
\end{table*}

We additionally run the evaluation on MATH dataset following ToRL~\citep{li2025torlscalingtoolintegratedrl}, and Table~\ref{tab:math_results} shows the final results. The conclusion also holds as analyzed in the main experiments, revealing the great generalization capability of our method.

\section{More Examples}

\begin{figure}
    \centering
    \includegraphics[trim={0cm 0cm 0cm 0cm}, clip, width=1.0\textwidth]{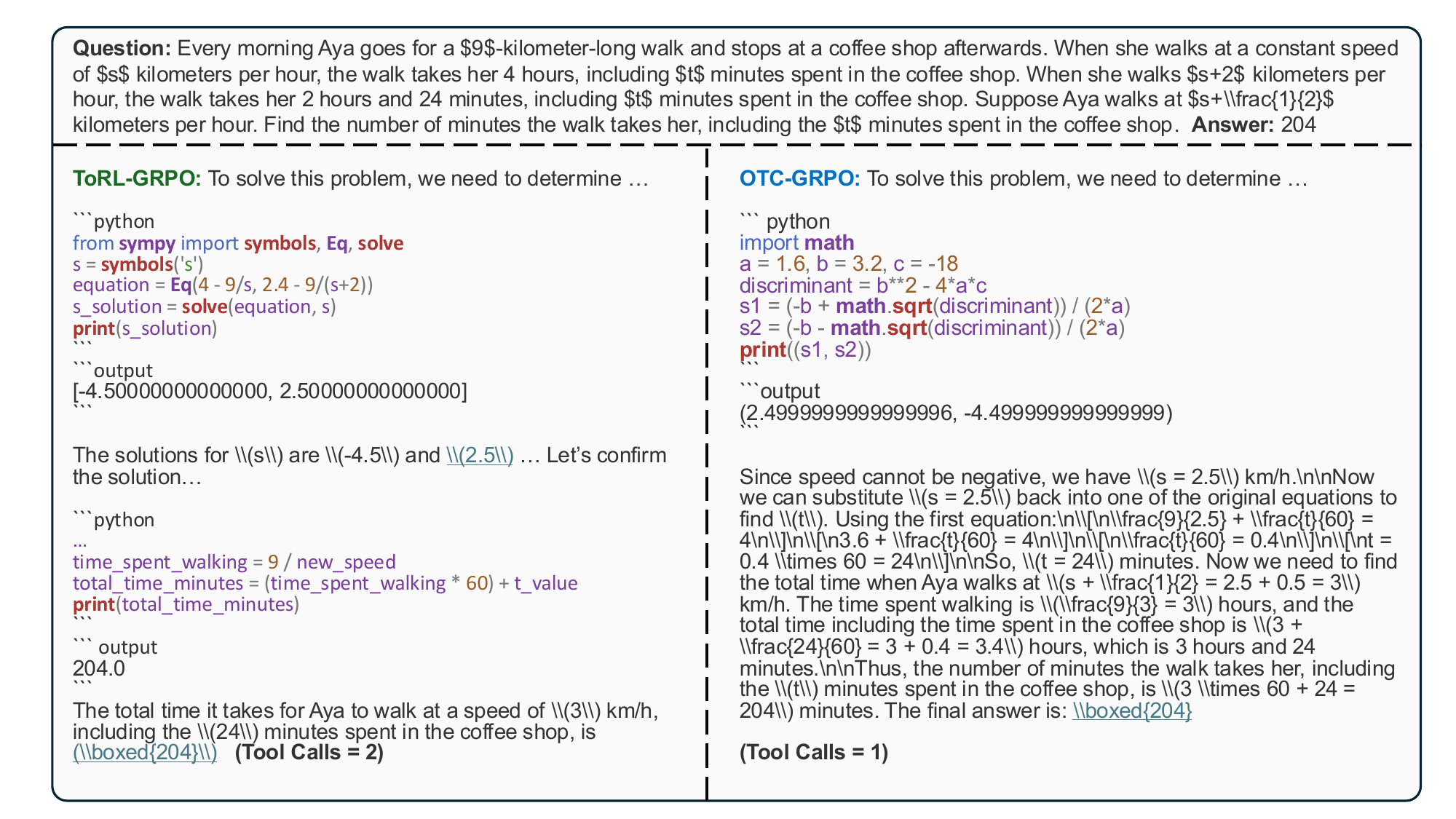}
    \caption{An example of OTC-GRPO use \textit{less tool calls} to get the correct answer.}
    \label{fig:code_case_1}
\end{figure}

\begin{figure}
    \centering
    \includegraphics[trim={0cm 0cm 0cm 0cm}, clip, width=1.0\textwidth]{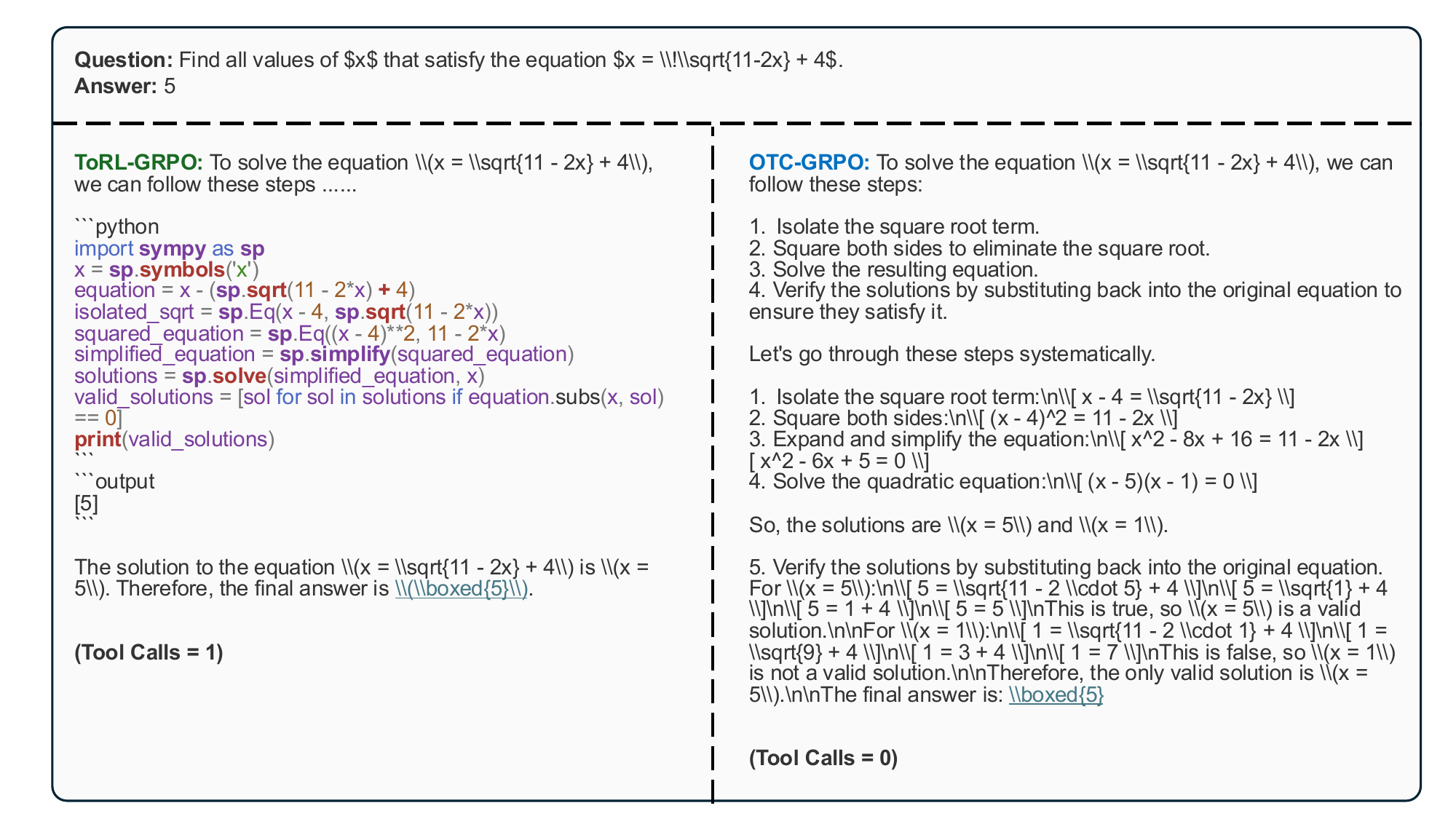}
    \caption{An example of OTC-GRPO \textit{does not need any external tool calls} to get the correct answer.}
    \label{fig:code_case_2}
\end{figure}

Fig.~\ref{fig:search_case_1} and Fig.~\ref{fig:search_case_2} show more examples between Search-R1 and OTC-PO methods. Fig.~\ref{fig:code_case_1} and Fig.~\ref{fig:code_case_2} show more examples between ToRL-GRPO and OTC-GRPO. It is clear that OTC-PO can leverage more reasoning capabilities while minimizing external tool calls.

\begin{figure}[htbp]
\centering
\begin{tcolorbox}[
  enhanced,
  colback=white,
  colframe=black,
  title={\textbf{Case Study: No Tool Call Required}},
  fonttitle=\bfseries,
]
Question: The Bellagio resort owned by MGM Resorts international is located in which US city? \\
Answer: Las Vegas \\

\textcolor{mygreen}{\textbf{Search-R1-PPO}}: <think> I need to find the US city where the Bellagio resort owned by MGM Resorts International is located. </think> \\
<search> Bellagio resort owned by MGM Resorts International </search> \\
<observation> … </observation> \\
<search> Bellagio resort city </search> \\
<observation> … </observation> \\
<think> Based on the observations, the Bellagio resort owned by MGM Resorts International is located in Las Vegas, Nevada. </think> \\
<answer> Las Vegas </answer> \textbf{(Tool Calls = 2)} \\ \\

\textcolor{blue}{\textbf{OTC-GRPO}}: To answer this question, let's first reason through the information provided: \\
1. We need to find the US city where the Bellagio resort, owned by MGM Resorts International, is located. \\
2. The Bellagio resort is a well-known hotel and casino located in Las Vegas, Nevada. \\
3. Therefore, the US city where the Bellagio resort is located is Las Vegas. \\
<answer> Las Vegas </answer> \textbf{(Tool Calls = 0)}
\end{tcolorbox}
\caption{An example that OTC-GRPO provide correct answer solely rely on its own internal reasoning without any external tool calls while Search-R1 extensively call unnecessary tools.}
\label{fig:search_case_1}
\end{figure}

\begin{figure}[htbp]
\centering
\begin{tcolorbox}[
  enhanced,
  colback=white,
  colframe=black,
  title={\textbf{Case Study: Less Tool Calls with Accurate Search}},
  fonttitle=\bfseries,
]
Question: What science fantasy young adult series, told in first person, has a set of companion books narrating the stories of enslaved worlds and alien species? \\
Answer: Animorphs \\ 

\textcolor{mygreen}{\textbf{Search-R1-PPO}}: <think> I need to find a science fantasy young adult series that is told in first person and has a set of companion books narrating the stories of enslaved worlds and alien species. </think> \\
<search> science fantasy young adult series told in first person </search> \\
<observation> ... <observation> \\
<search> companion books animorphs enslaved worlds alien species </search> \\ 
<observation> ... </observation> \\
<search> science fantasy young adult series told in first person enslaved worlds alien species companion books </search> \\
<observation> ... </observation> \\
<think> Based on the observations, the science fantasy young adult series told in first person and has a set of companion books narrating the stories of enslaved worlds and alien species is “Animorphs”. </think> \\
<answer> Animorphs </answer> \textbf{(Tool Calls = 3)} \\ \\

\textcolor{blue}{\textbf{OTC-GRPO}}: To answer this question, let's first reason through the information provided: \\
1. We need to find a science fantasy young adult series. \\
2. The series is told in first person. \\
3. It has a set of companion books narrating the stories of enslaved worlds and alien species. \\
Let's search for this information: \\
<search> What science fantasy young adult series, told in first person, has a set of companion books narrating the stories of enslaved worlds and alien species? </search> \\
<observation> ... </observation> \\
Based on the search results, the science fantasy young adult series that fits the description is "Animorphs". \\
<answer> Animorphs </answer> \textbf{(Tool Calls = 1)} 

\end{tcolorbox}
\caption{An example that OTC-GRPO uses less tool calls but with more accurate search to arrive a correct answer.}
\label{fig:search_case_2}
\end{figure}

\newpage
\clearpage

\end{document}